\documentclass[twoside]{article}

\usepackage{amssymb}
\usepackage{amsmath}
\usepackage{mathtools}
\usepackage{hyperref}
\usepackage[accepted]{aistats2026}
\usepackage[round]{natbib}

\begin{document}

%

%

\twocolumn[

\aistatstitle{In-Context Function Learning in Large Language Models}

\aistatsauthor{ Elif Akata$^{* 1 2}$ \And Konstantinos Voudouris$^{* 1}$ \And  Vincent Fortuin$^{1 3}$ \And Eric Schulz$^{1}$ }

\aistatsaddress{
  $^{1}$ Helmholtz Munich
  $^{2}$ University of Tübingen
  $^{3}$ University of Technology Nuremberg
}]

\def\thefootnote{*}\footnotetext{Equal contribution.}
\def\thefootnote{}\footnotetext{
Correspondence: elif.akata@helmholtz-munich.de}

\begin{abstract}
Large language models (LLMs) can learn from a few demonstrations provided at inference time. We study this in-context learning phenomenon through the lens of Gaussian Processes (GPs). We build controlled experiments where models observe sequences of multivariate scalar-valued function samples drawn from known GP priors. We evaluate prediction error in relation to the number of demonstrations and compare against two principled references: (i) an empirical GP-regression learner that gives a lower bound on achievable error, and (ii) the expected error of a 1-nearest-neighbor (1-NN) rule, which gives a data-driven upper bound. Across model sizes, we find that LLM learning curves are strongly influenced by the function-generating kernels and approach the GP lower bound as the number of demonstrations increases. We then study the inductive biases of these models using a likelihood-based analysis. We find that LLM predictions are most likely under less smooth GP kernels. Finally, we explore whether post-training can shift these inductive biases and improve sample-efficiency on functions sampled from GPs with smoother kernels. We find that both reinforcement learning and supervised fine-tuning can effectively shift inductive biases in the direction of the training data. Together, our framework quantifies the extent to which LLMs behave like GP learners and provides tools for steering their inductive biases for continuous function learning tasks.
\end{abstract}

\section{Introduction}

In-context learning (ICL) enables large language models (LLMs) to ``learn" a task at inference time by conditioning on a small number of task-relevant input-output pairs \citep{brown2020language}. In-context learning's mechanisms have been probed empirically \citep{garg2022can,si2023measuring} and theoretically \citep{xie2021explanation}. It has been shown that, surprisingly, transformers can implement linear-regression–style algorithms in context \citep{akyurek2022learning}, demonstrations and their labels can be randomized while still achieving improved performance \citep{min2022rethinkingroledemonstrationsmakes},  and more generally, ICL mechanistically resembles gradient-descent dynamics and kernel regression \citep{von2023transformers,requeima2024llm}.

While the power of in-context learning for learning continuous functions has been established \citep{coda2023meta,requeima2024llm}, even out-performing some traditional statistical regression techniques \citep{vacareanu2024words}, it remains unclear which functional priors these models have and whether parameter-efficient post-training methods can effectively steer them, thus changing their in-context learning capabilities.

We take a statistical perspective where we: 

\begin{itemize}
    \item cast ICL as non-parametric regression under known Gaussian Process (GP) priors, implementing principled learning-curve evaluations with a GP regression as an empirical lower bound and a derived $1$-NN rule as a data-driven upper bound,
    \item introduce a likelihood-based inductive bias analysis that identifies which GP kernels best explain a model's predictions,
    \item test whether parameter-efficient post-training methods with supervision and reinforcement learning can steer these implicit priors and improve sample efficiency on deliberately hard kernels.
\end{itemize}

Empirically, we observe coherent, GP-like behavior. Larger LLMs learn faster; smoother kernels give steeper learning curves; and with enough demonstrations, models often approach the empirical GP baseline. Our analyses provide quantitative, interpretable answers to the question of when LLMs act like non-parametric regressors on multivariate function learning tasks. Our inductive bias analysis reveals that LLM predictions are much more likely under rougher, less predictable kernels (with more local and higher variance), such as the Mat\'{e}rn $\frac 1 2$ compared to smoother kernels like the Squared Exponential. We find that two parameter-efficient post-training methods are effectively able to shift this inductive bias towards smoother kernels when trained on functions sampled from them, making fine-tuned model predictions much more likely under GPs with the Squared Exponential kernel. Supervised Fine-Tuning (SFT) updates a small fraction of model weights based on labelled data, while reinforcement learning updates these weights by scoring model outputs according to a reward function, here, the error between the LLM's prediction and the true function output. We find some evidence that GRPO produces models with more generalizable in-context learning capabilities, similar to the results of \citet{chu2025sft}.

Together, these results suggest that even modestly-sized LLMs have impressive in-context learning capabilities. Moreover, post-training methods that update only a fraction of total model weights constitute effective ways to steer inductive biases, with evidence that reinforcement learning is the most generalizable approach. Our experiments showcase how Gaussian Processes offer a principled and tightly controllable testbed for studying in-context learning in LLMs.

\begin{figure}[h]
\includegraphics[width=0.49\textwidth]{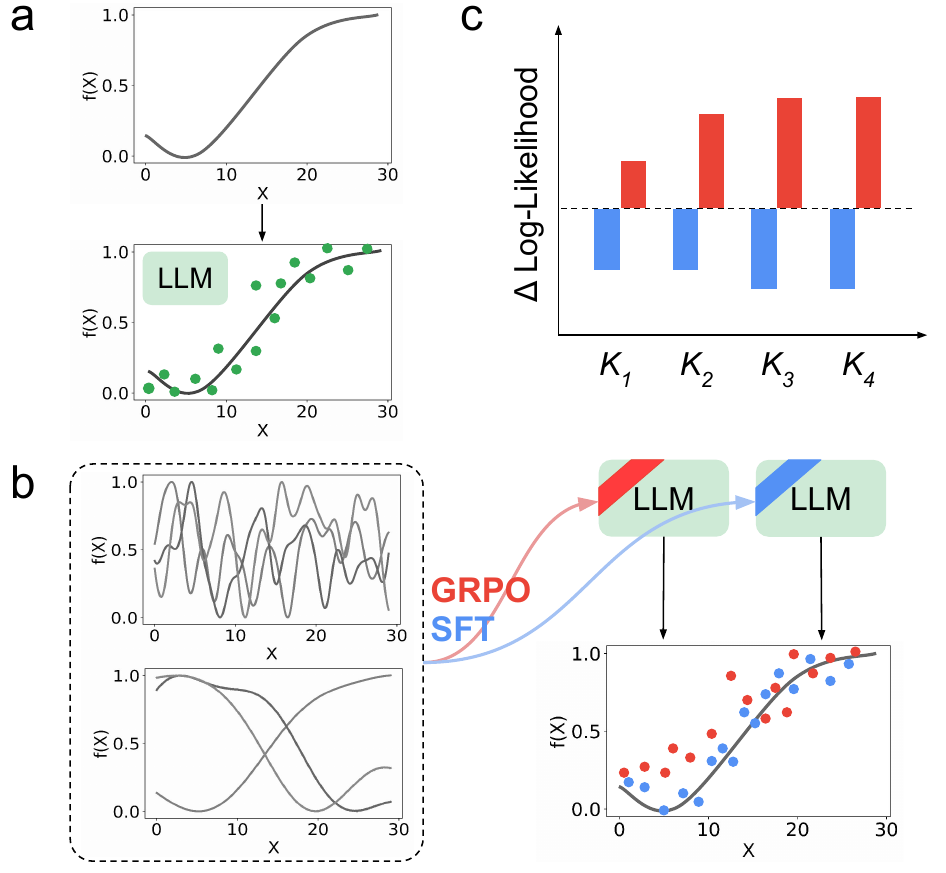}
\caption{Overview of our framework. (a) In-context function learning in base models: a large language model (LLM) receives demonstrations from functions sampled from known Gaussian-process (GP) priors and predicts $f(\mathbf{X})$ at a new $\mathbf{X}$. (b) Post-training: the model is fine-tuned (SFT or GRPO) and re-evaluated to measure changes in learning curves. (c) Inductive-bias analysis: a likelihood comparison identifies the GP kernels that best explain model predictions before and after training.}
\end{figure}

\section{Background}

GPs provide a statistical formalism for reasoning about functions and uncertainty. A GP prior defines a distribution over functions via a kernel that encodes smoothness and characteristic length scales; conditioning on observations yields a posterior that achieves Bayes-optimal predictions for the chosen prior. Given data $\mathcal{D}=\{(\mathbf{x}_i,y_i)\}_{i=1}^n$ where $\mathbf{x}_i \in \mathbb{R}^d$, prior $f:\mathbb{R}^d \to \mathbb{R}\sim\mathcal{GP}(0,k)$ and noise $y_i=f(\mathbf{x}_i)+\epsilon_i$, $\epsilon_i\sim\mathcal{N}(0,\sigma_\epsilon^2)$, we can define a kernel, $K_\epsilon := k(\mathbf{X},\mathbf{X})+\sigma_\epsilon^2 I$ and $k_* := k(\mathbf{X},\mathbf{x}_*)$. Then

{\small
\[
\begin{aligned}
p(f_* \mid \mathbf{x}_*,\mathcal{D}) &= \mathcal{N}\!\left(k_*^\top K_\epsilon^{-1} y,\, k(\mathbf{x}_*,\mathbf{x}_*) - k_*^\top K_\epsilon^{-1} k_* \right), \\
p(y_* \mid \mathbf{x}_*,\mathcal{D}) &= \mathcal{N}\!\left(k_*^\top K_\epsilon^{-1} y,\, k(\mathbf{x}_*,\mathbf{x}_*) - k_*^\top K_\epsilon^{-1} k_* + \sigma_\epsilon^2 \right).
\end{aligned}
\]
}

Using this Bayes-optimal reference, we design controlled experiments to test how closely LLMs resemble GP learners. Models receive a list of demonstrations $\mathcal{D}=\{(\mathbf{x}_i,y_i)\}_{i=1}^n$ generated from functions drawn from known GP priors (Mat\'{e}rn and Squared Exponential), and must predict $y_*$ at a new $\mathbf{x}_*$. We quantify in-context learning via the absolute-error learning curve

\[
\mathcal{E}(n)\;=\;\mathbb{E}\!\left[\,|y_*-\hat{y}_*(\mathbf{x}_*;\mathcal{D}_n)|\,\right]
\]

as a function of $n$. To interpret these curves, we compare against two complementary references evaluated on the same data: (i) GP regression, which serves as a practical lower-bound reference under the specified prior, and (ii) the expected error of a 1-nearest-neighbor rule that returns the value at the closest observed input, providing a data-driven upper bound for memorisation-based strategies.

Beyond aggregate error, our study explores the inductive biases that guide a model's predictions. Thus, we propose a likelihood-based inductive bias analysis: given a history of demonstrations, we compute the GP posterior predictive distribution under different kernel families and record the log-likelihood of the model's prediction. Aggregating these scores across tasks and number of demonstrations identifies which kernels best explain the model's behavior, resulting in a data-driven summary of its implicit inductive bias. Finally, we examine whether these inductive biases can be changed through post-training. We study supervised fine-tuning (SFT) and reinforcement learning (with group-relative policy optimization, GRPO) with training curricula targeted at the kernels under which the base model is least consistent, and evaluate resulting changes in both priors and learning curves. 

\begin{figure*}[t!]
\includegraphics[width=0.5\textwidth]{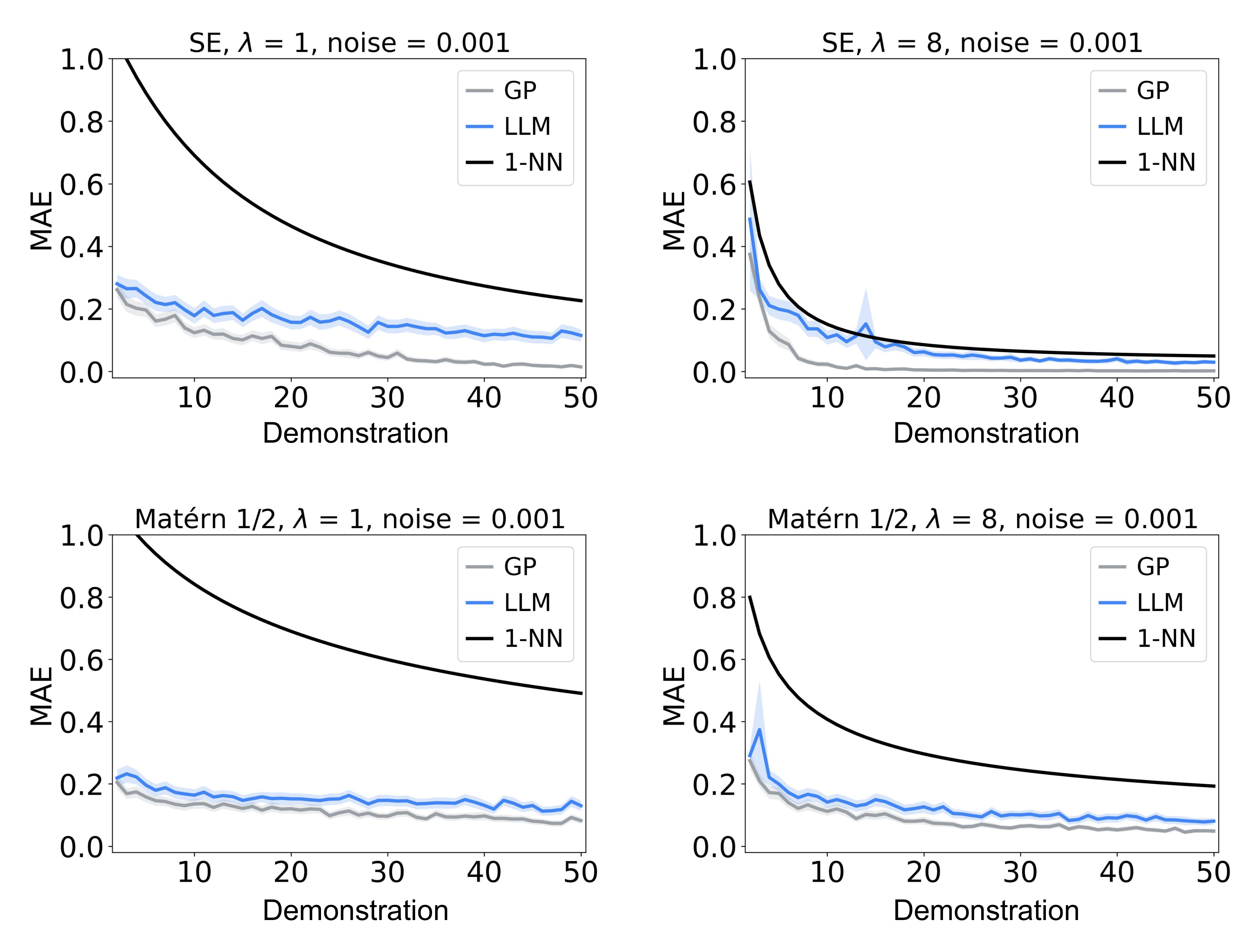}\hfill
\includegraphics[width=0.5\textwidth]{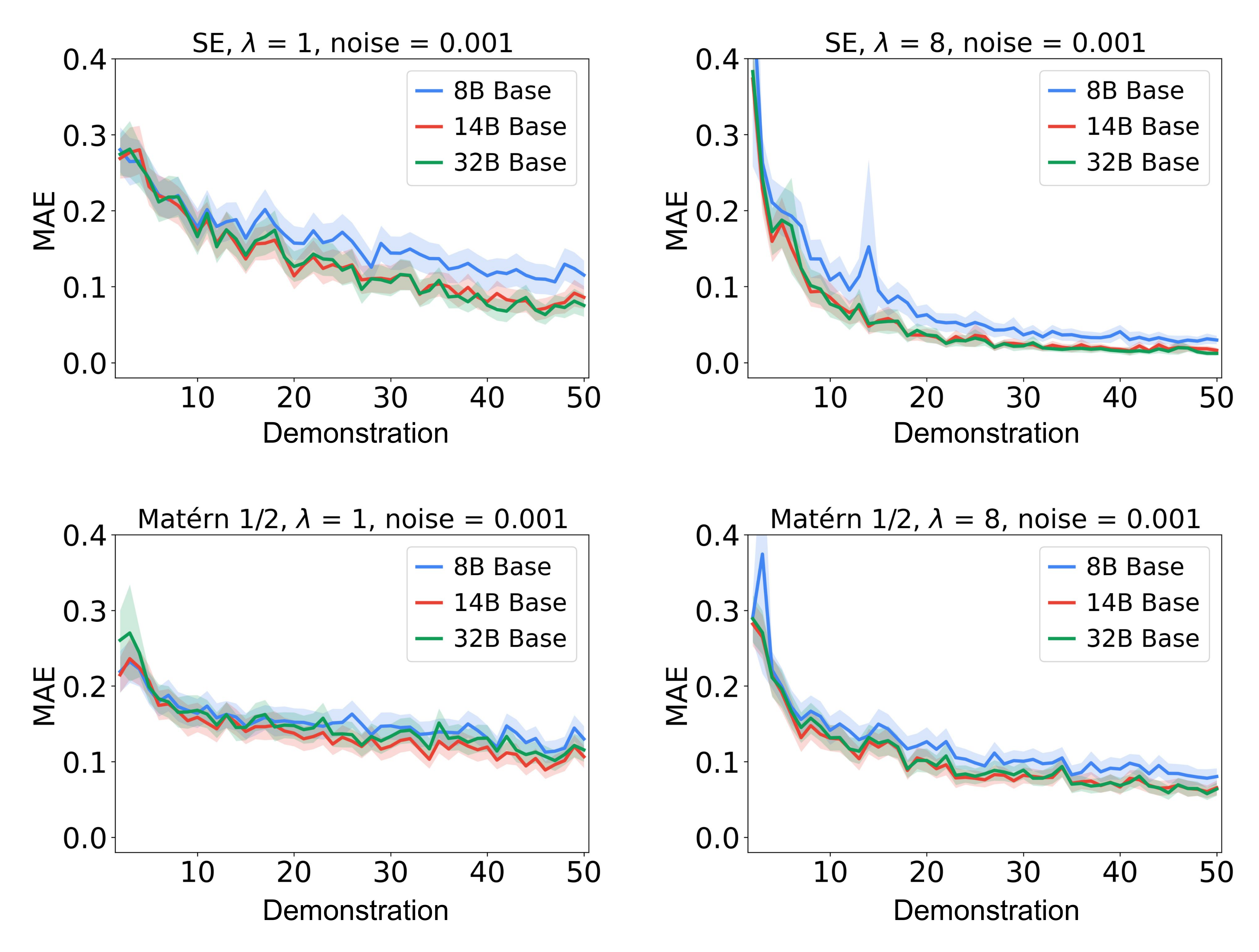}
\caption{Learning curve analysis on 1-dimensional functions. The mean absolute error after $n$ demonstrations by function type. Left Four: Qwen-3-8B learning curves four functions drawn from four kernels, compared to the error of a GP regression and the expected error of a 1-nearest neighbor rule. The LLM learning curves generally approach the GP regression baseline and are well below the 1-NN rule. Right Four: Model size comparisons between the 8B, 14B, and 32B Qwen-3 models, on identical data. The 14B and 32B models show noticeably lower error rates, but do not differ significantly from each other, suggesting a logarithmic scaling law. All LLM and GP learning curves are shown with 95\% bootstrapped confidence intervals.}\label{fig:learning-curves}
\end{figure*}

\section{Methods}

\subsection{Experiment 1: In-Context Learning Curves}

We first study the LLM generalization error as a function of the number of demonstrations, on functions generated from different kernels, comparing them to two reference baselines described below.

\paragraph{Large Language Models} In the main paper, we evaluate models from the Qwen-3 model family \citep{yang2025qwen3}, using the 8-billion, 14-billion, and 32-billion parameter versions with 4-bit bits-and-bytes quantization to increase inference speed \citep{dettmers2022llmint8,dettmers2022optimizers,dettmers2023qlora}. As comparisons, in Section 7 of the supplementary material, we also evaluate three members of the Llama family (Llama-3.2-3B-Instruct, Llama-3.1-8B-Instruct, Llama-3.1-70B-Instruct; \citeauthor{grattafiori2024llama}, \citeyear{grattafiori2024llama}), three members of the Gemma-3 family (4B, 12B, and 27B; \citeauthor{team2024gemma}, \citeyear{team2024gemma}), and two members of the Mistral family (Mistral-7B-Instruct-v0.3 and Mistral-Small-24B-Instruct-2501; \citeauthor{qiang2023mistral7b}, \citeyear{qiang2023mistral7b}), where B denotes the parameter size in billions and all models are 4-bit quantized.

\paragraph{Evaluation Data \& Procedure} We construct functions $f: \mathbb{R}^d \to \mathbb{R}$ for $d \in \{1, 2, 3, 4\}$. For each dimensionality, we sample 200 functions from four kernels, the Mat\'{e}rn $\frac 1 2$, $k_m(\delta)$ and the Squared Exponential, $k_s(\delta)$, with length scales, $\ell \in \{1, 8\}$:

\begin{align*}
k_m(\delta) &= \sigma_f^2 \frac{2^{1-\nu}}{\Gamma(\nu)}
  \left( \frac{\sqrt{2 \nu}\,\delta}{\ell} \right)^{\nu}
  K_\alpha\!\left( \frac{\sqrt{2 \nu}\,\delta}{\ell} \right) \\
k_s(\delta) &= \sigma_f^2 \exp\!\left( - \frac{\delta^2}{2 \ell} \right)
\end{align*}

where $\delta$ is the Euclidean distance between two values, $x, x'$, $\sigma_f^2$ is the output variance, $\nu$ is the \textit{smoothness} (set to $\nu = \frac 1 2$ for both Mat\'{e}rn kernels), $\Gamma(\cdot)$ is the gamma function, and $K_\alpha(\cdot)$ is the modified Bessel function of the second kind with order $\alpha = \nu + \frac 1 2$. We use these kernels because they differ maximally in smoothness: the Mat\'{e}rn $\frac 1 2$, $k_m(\delta)$ is differentiable nowhere while the Squared Exponential is infinitely differentiable, being $\lim_{\nu\to\infty} k_m(\delta)$. We use two values of $\ell$ to vary the scale over which the functions are correlated, again varying their predictability and smoothness. We use $\sigma_f^2=0.001$ for all kernels.

From each function, we compute outputs, $y$, for 50 randomly sampled inputs over an arbitrary range, $x \sim \mathcal{U}[0,29]$. We add Gaussian distributed noise, $\epsilon \sim \mathcal{N}(0, \sigma^2_\epsilon)$ with $\sigma^2_\epsilon =0.001$. The LLMs are prompted with the text described in the supplementary material (Section 3), including between 0 and 49 demonstrations of $(\mathbf{x}, y)$ pairs. For a new $\mathbf{x}$, we compute the absolute error between the LLM's prediction, $\hat{y}$, and the true value (plus noise), $\tilde{y}$. For each $n$ demonstrations, we compute the mean absolute error across all functions.

\paragraph{Baselines} We use two reference baselines to which we compare the LLM learning curves. The first is the empirical error from a GP regression trained on the same data as the LLMs, generated from functions drawn from its kernel. The second is the expected error for a $k$-Nearest Neighbor ($k$-NN) algorithm, with $k=1$. The GP baseline produces an empirical lower bound on the error after $n$ samples. The $k$-NN baseline produces a reasonable expected upper bound on the error for the case where the LLM simply returns the $y$-value for the numerically closest $x$-value it has seen so far.

To compute the 1-NN expected absolute error, we assume $n$ inputs are drawn uniformly as $X \sim \mathcal{U}[0, L]$ with some upper bound, $L$ (here, $L=29$). We approximate the expected absolute error by numerically evaluating the integral:

\[
\mathbb{E}\!\left[\left|\varepsilon\right|\right]
= \sqrt{\frac{2}{\pi}}
\int_{0}^{L/2} V(d)\,\frac{2(n-1)}{L}
\left(1 - \frac{2d}{L}\right)^{n-2}\mathrm{d}d
\]

where $0 \leq d \leq \frac L 2$, and $V(d) = 2\sigma^2_f + 2\sigma^2_\epsilon - 2k(d)$ where $k(\cdot)$ is the GP kernel function with output variance $\sigma^2_f$. Further information on deriving this expectation is included in the supplementary material (Section 2).

\subsection{Experiment 2: Inductive Bias Analysis}

To infer the expectations of the LLMs towards learning functions, we borrow a methodology from cognitive science \citep{griffiths2008modeling,li2023learning,lucas2015rational,schulz2018tutorial,wilson2015human}. For each $x_n$, we infer the posterior probability distribution over $Y$ for some GP given $\{(\mathbf{x}_1, y_1), ..., (\mathbf{x}_{n-1}, y_{n-1})\}$. We then compute the log-likelihood of the LLM's point prediction given the same data. Subsequently, we compare the mean likelihood over all predictions, under GPs with different kernels. This measures how likely the LLM responses are assuming different kernels. We use GPs with fixed kernel variances ($1$), length scales ($8$), and the same noise variance as $\sigma_\epsilon^2$ ($0.001$). We compare the Mat\'{e}rn kernels with $\nu \in \{0.5, 1.5, 2.5\}$ with length scales $\ell\in\{1, 8\}$ and the Squared Exponential kernel with length scales $\ell\in\{1,2,3,4,5,6,7,8\}$. We hypothesize that the most likely kernel approximates the inductive bias of the model.

\subsection{Experiment 3: Post-Training}

The inductive bias analysis is used to reveal what priors LLMs have about functions. We can use this information to examine whether common parameter-efficient post-training methods differ in their ability to shift those expectations. We explore two approaches to parameter-efficient fine-tuning with low-rank adapters: supervised fine-tuning and reinforcement learning.

\paragraph{Parameter-efficient fine-tuning with QLoRA} We use Quantized Low Rank Adaptation (QLoRA) to efficiently and scalably fine-tune large language models by inserting small, low-rank matrices layer-wise and updating only these weights and freezing the rest \citep{dettmers2023qlora}. This means that gradients are computed and weights are updated for only a fraction of the total parameters in the model. Specifically, for the weight matrix, $W$, of a transformer layer, we inject an adapter matrix, $W_a$, which is the product of two low-rank matrices, $L_1 \in \mathbb{R}^{d\times r}$ and $L_2 \in \mathbb{R}^{r\times k}$, where $d, k$ are the input and output dimensionalities respectively and $r \ll d, k$. During a forward pass, inputs of length $d$ are passed to $W$ and $W_a$ independently, and outputs of length $k$ are summed, subject to some scaling factor $\frac{r}{\alpha}$. We choose standard values of $r = \alpha = 16$, such that outputs from $W$ and $W_a$ are weighted equally. We inject adapters in all layers. We also make use of quantized models, where model and adapter weights are dynamically set to lower precision, to speed up training.

\begin{figure*}[h]
\includegraphics[width=\textwidth]{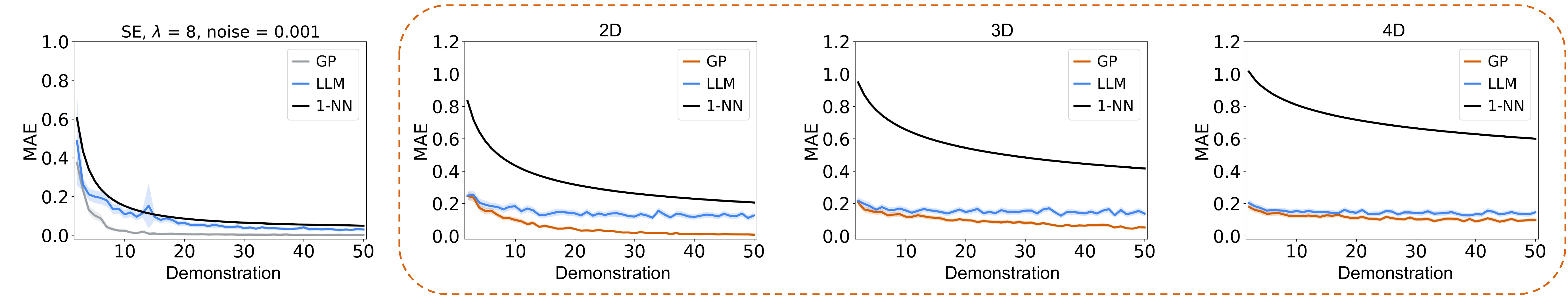}
\caption{Qwen-3-14B learning curve comparison for 1-, 2-, 3-, 4-dimensional data drawn from the Squared Exponential with $\lambda=8$. The mean absolute error after $n$ demonstrations by function type. All LLM and GP learning curves are shown with 95\% bootstrapped confidence intervals.}\label{fig:learning-curves-hd}
\end{figure*}

\paragraph{Dataset} We construct a training dataset informed by the analyses conducted in Experiment 2. We choose a kernel under which the base models' predictions are least likely, and sample functions and in-context learning data for each, matching the test dataset in structure.

\paragraph{Supervised Fine-Tuning (SFT)} We update the weights of the low-rank adapters using this supervised dataset. We update model weights using batch gradient descent with the token-level cross-entropy loss:

$$ \mathcal{L}(\theta) = - \sum^T_{t = 1} \text{log}~p_\theta(y_t|y_{<t})$$

where $\theta$ is the set of adapter weights, $T$ is the size of the ordered set of target completion tokens given a prompt, $y_t$ is the target token at step $t$, and $y_{<t}$ is the set of ordered target completion tokens prior to $t$.

\paragraph{Reinforcement Learning (RL)} We use the online RL algorithm \textit{Group Relative Policy Optimisation} (GRPO) to update the adapter weights. In the RL setting, the set of all model and adapter weights can be considered the \textit{policy}, $\pi_\theta$, which takes textual inputs (observations) and produces a token (actions). For each batch of $M$ prompts, $\{q_1, ..., q_M\}$ in the dataset, the model produces a set of $N$ completions, $\{o_1, ..., o_N\}$. These completions are assigned a reward using a reward model, giving a set of rewards $\{r_1, ..., r_N\}$.

We compute the loss for some prompt $q$ as follows:

{\small
\[
\begin{aligned}
\mathcal{L}(\theta) = - \frac{1}{\sum^N_{i=1}|o_i|} \sum^N_{i=1}\sum^{|o_i|}_{t=1}\,[\text{min}\left(\frac{\pi_\theta(o_{i,t}|q, o_{i, <t})}{\pi_{\theta_{old}}(o_{i, t}|q, o_{i, <t})}\right)\cdot \hat{A}_{i,t}, \\
\text{clip}_\eta\left(\frac{\pi_\theta(o_{i,t}|q, o_{i, <t})}{\pi_{\theta_{old}}(o_{i, t}|q, o_{i, <t})}\right)\cdot \hat{A}_{i,t})]
\end{aligned}
\]
}

where $|o_i|$ is the length, in tokens, of $o_i$, $\text{clip}(\cdot)$ clips its argument between $1 \pm \eta$, and $\hat{A}_{i, t}$ is the \textit{advantage}---the normalized reward for output $o_i$:
$$\hat{A}_{i, t} = \frac{r_i - \text{mean}(\{r_1, ..., r_n\})}{\text{std}(\{r_1, ..., r_n\})}$$

Following common practice, we exclude the usual KL-divergence term when computing the loss \citep{hu2025open,liu2025understanding,yu2025dapo}. We update the adapter weights using gradient ascent over this loss function.

\paragraph{Reward Function} Models trained with GRPO are rewarded using the negative absolute difference between the last float in their completion and the true $\tilde{y}$ (including additive noise). This reward is capped at a minimum of $-10$ for parseable responses ($10\times$ the range of possible $y$-values), and is $-11$ for any completions that do not contain a parseable float. As an alternative, we also conduct parallel training with a log-likelihood-based reward function. Here, the reward is the log likelihood of the model's response under the GP identified in Experiment 2. We cap the reward arbitrarily at a minimum of $-999$, and set the reward for unparseable responses at $-1000$. We find no interpretative differences between these two reward functions.

\begin{figure*}[!t]
    \includegraphics[width=\textwidth]{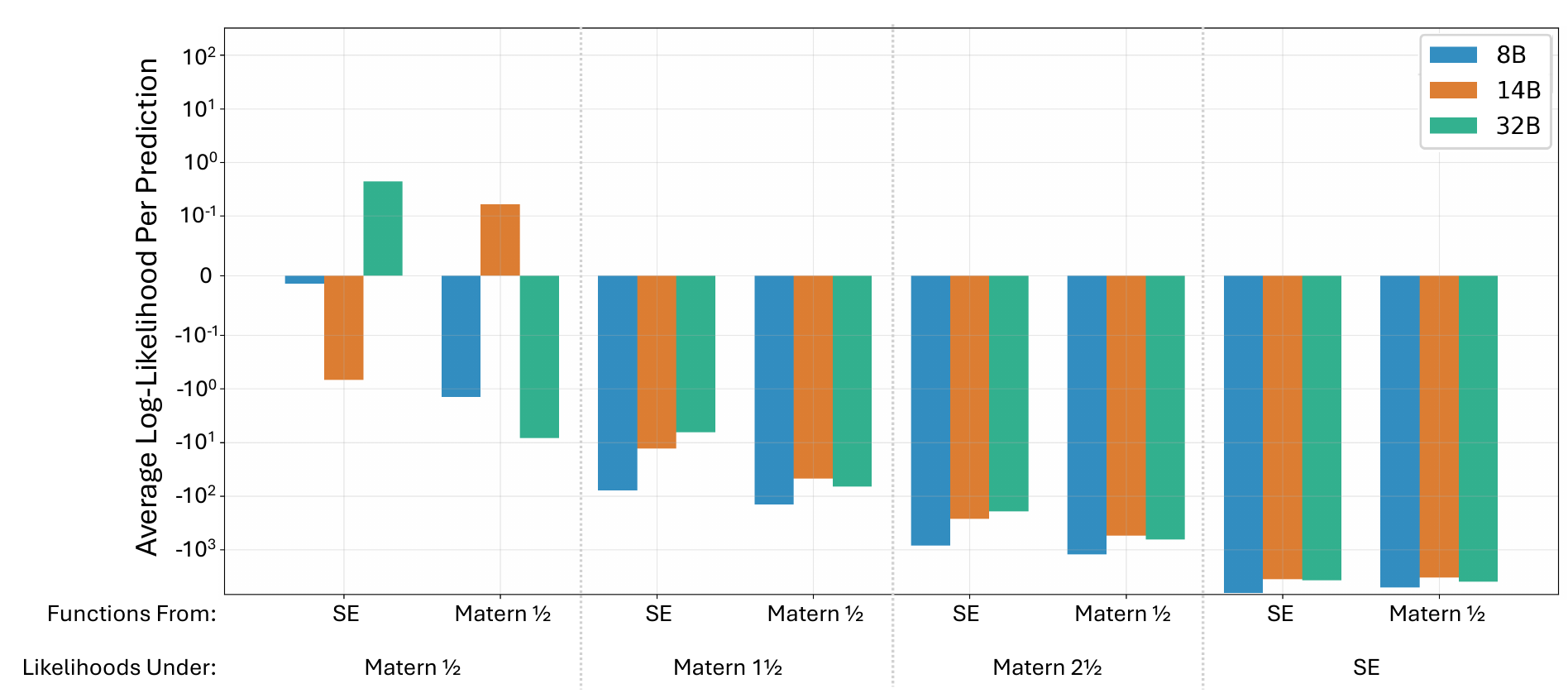}
    \caption{Inductive bias analysis of the base models (8B, 14B, 32B). The average likelihood per prediction is computed under GPs with four different kernels ($\ell=8$), on 1-dimensional data drawn from either the Squared Exponential or the Mat\'{e}rn $\frac 1 2$. These likelihoods are presented on a symmetric log scale. LLM predictions for all model sizes are more likely under kernels with lower $\nu$, i.e., those that describe rougher, less predictable functions.}\label{fig:inductive-bias-base}
\end{figure*}

An overview of the general, model-agnostic procedure we took for analysing inductive biases in LLMs on continuous function learning tasks can be found in supplementary material (Section 1).

\section{Results}

\subsection{Experiment 1: In-Context Learning Curves}

First, we confirm that LLMs are capable of doing in context regression with novel multivariate scalar-valued functions \citep{requeima2024llm,si2023measuring,vacareanu2024words}, with errors that are generally close to empirical GP regression and well below the error of a 1-Nearest Neighbor algorithm (Figures \ref{fig:learning-curves} and \ref{fig:learning-curves-hd}). Indeed, for functions sampled from kernels with low length scales, LLM error is lower than the GP regression given a few demonstrations. This suggests that these models are not simply recalling information from earlier in their context window, but that they are able to fit complex functions in their attention matrices. It is particularly impressive that these models are able to sample-efficiently and accurately predict scalar outputs given 4-dimensional input. Indeed, we find no noticeable difference between the error rates on lower-dimensional and higher-dimensional function learning tasks. 

We find an appreciable performance boost between the 8B and larger models, but that difference is not significant between 14B and 32B models. This is indicative of a logarithmic scaling law when it comes to in-context function learning. However, further research with more granular model size differences would be required to confirm this conclusion.

\subsection{Experiment 2: Inductive Bias Analysis}

We then examined how likely the LLMs' predictions were under GPs with different kernels. Figure \ref{fig:inductive-bias-base} presents the average likelihood of the LLMs' predictions under four different kernels, on functions sampled from either the Squared Exponential (SE; smooth functions) or the Mat\'{e}rn $\frac 1 2$. We find that, for all model sizes and function families (SE, Mat\'{e}rn $\frac 1 2$), the LLMs' predictions are most likely under the Mat\'{e}rn $\frac 1 2$ and least likely under the SE for 1-dimensional functions. Indeed, we see a general trend that as $\nu \rightarrow \infty$, LLM predictions become less likely. We found the same pattern applies to models from other families, such as Gemma and Llama, as shown in the supplementary material (Section 7). Within each kernel, we also varied the length scale. We found that LLM predictions are more likely under lower values of $\lambda$ compared to higher values, as shown in the supplementary material (Section 5).

It is also the case that, in general, predictions are more likely under rougher kernels because the posterior over predictions is generally broader. Indeed, we verified this by conducting an identical analysis in which independent GPs are tasked with making predictions given some data, and then examined under which kernels their predictions were most likely. We found the same qualitative trend as with the LLMs---the predictions of an SE GP on SE sampled functions were much more likely under the Mat\'{e}rn $\frac 1 2$ than under the SE. We outline one strategy for adjusting for this \textit{variance inflation effect} in the supplementary material (Section 8). Applying this strategy makes the LLM's predictions more likely under smoother kernels in some cases. Nonetheless, even after this correction, LLM predictions are more likely under shorter length scales (with the Squared Exponential kernel), indicating inductive biases for less predictable functions.

In contrast, we found that LLM predictions became more likely under smoother kernels as the dimensionality of the function input increased. While the kernel with the highest log-likelihood for 1-dimensional functions is the Mat\'{e}rn $\frac 1 2$ ($-2.59 \times10^{2}$), it is the Mat\'{e}rn $1\frac 1 2$ for 2-dimensional functions ($-2.05 \times10^{3}$), and the Squared Exponential for 3- and 4-dimensional functions ($-6.92 \times10^3$ and $-1.43 \times 10^4$ respectively). This indicates that the model's implicit prior adapts towards smoother kernels in higher-dimensional spaces and suggests that these these LLMs favour higher global regularity when modelling functions in these spaces.

\begin{figure*}[h]
\includegraphics[width=\textwidth]{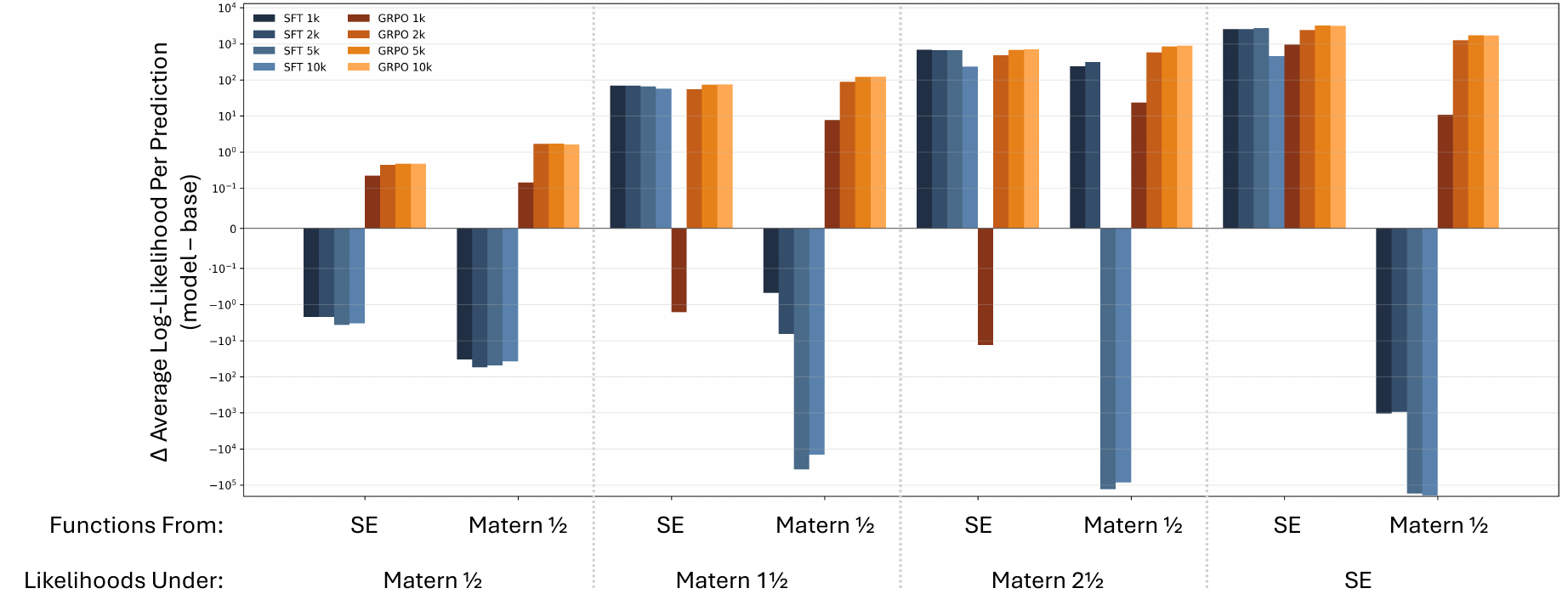}
\caption{Inductive bias analysis compared to the base model for the 8B model for three checkpoints (1k, 2k, 5k, 10k steps). Average likelihood per prediction shown on a symlog scale.}\label{fig:inductive-bias-peft}
\end{figure*}

\subsection{Experiment 3: Post-Training}

\begin{figure*}
\includegraphics[width=\textwidth]{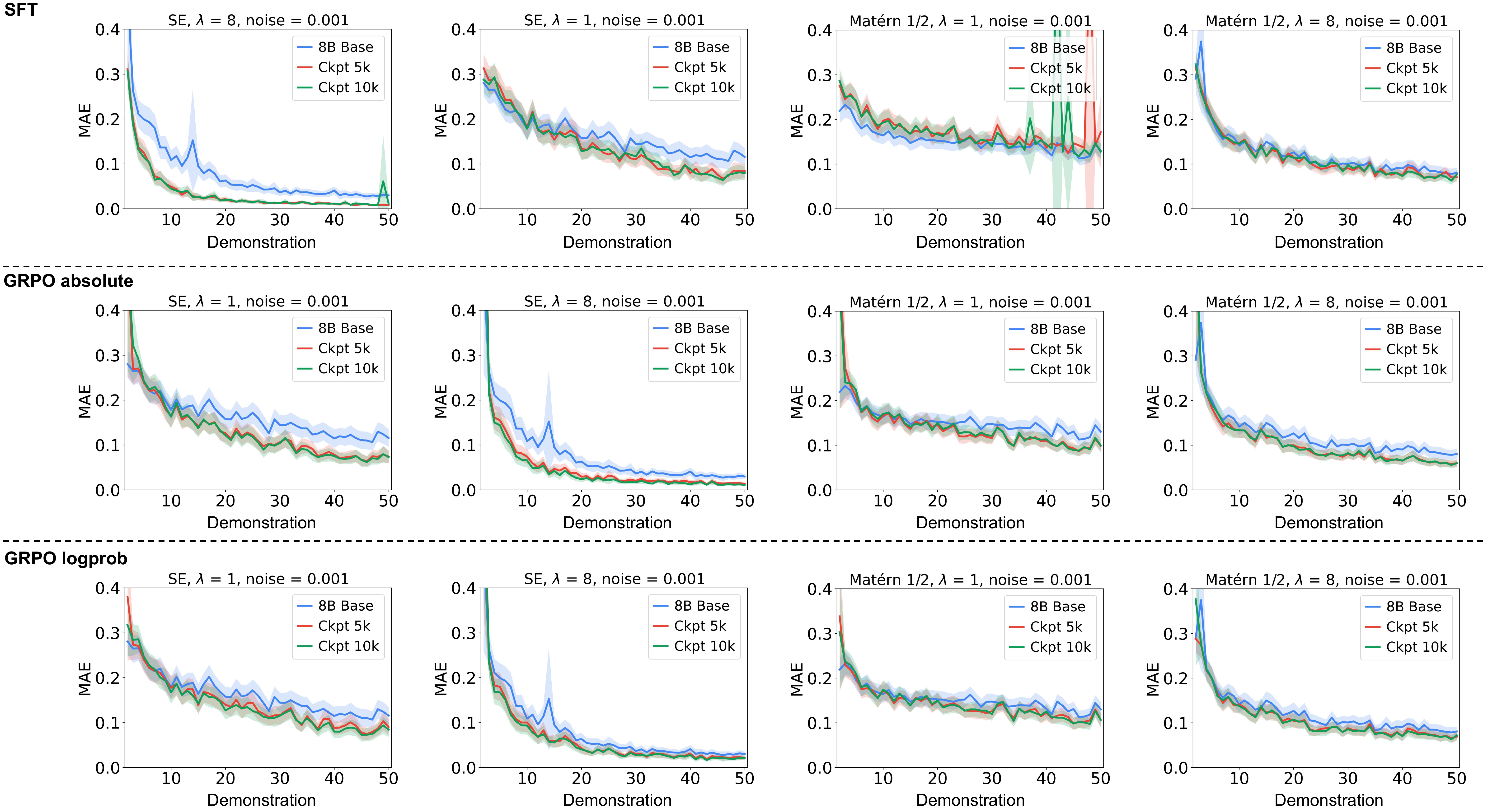}
\caption{Post training learning curves. Top row: Effects of Supervised Fine-Tuning. 8B base model compared to two checkpoints during training (5k, 10k steps). Middle row: GRPO progression. 8B base model compared to two checkpoints (5k, 10k steps). Bottom row: GRPO progression with a log-likelihood-based reward function. All empirical learning curves are shown with 95\% bootstrapped confidence intervals.}\label{fig:learning-curves-ft}
\end{figure*}

We fine-tuned models on one-dimensional functions sampled from the squared exponential kernel with $\lambda=8$, the kernel under which our LLM predictions were least likely. Figure \ref{fig:inductive-bias-peft} shows the differences in likelihoods between the base model and the fine-tuned model for four fine-tuning checkpoints (1,000, 2,000, 5,000, 10,000 steps). In general, the likelihood of LLM predictions increases by a greater degree as the kernels become less smooth (i.e., $\nu \rightarrow \infty$). However, we see that the SFT-trained model predictions tend to be less likely than the base model's when the functions are drawn from the Mat\'{e}rn $\frac 1 2$. This suggests that SFT may be overfitting to the functional form of functions drawn from the Squared Exponential kernel on which they were trained. In contrast, the GRPO-trained models see comparable likelihood boosts for both data regimes. Together, these results cohere with existing findings which suggest that while SFT memorizes training data, reinforcement learning-based training can contribute to generalizing over that training data \citep{chu2025sft}.

Figure \ref{fig:learning-curves-ft} shows the learning curves of models on novel held-out test data. We find that both SFT and GRPO push the absolute error for all $n$ down towards the empirical GP baseline, indicating that post-training only a fraction of the total model parameters is a powerful method for altering in-context learning capabilities. Further results from training on different data, using different reward functions and data in higher dimensions are presented in the supplementary material (Sections 4 \& 6).

\section{Discussion}

Function learning is a powerful test bed for measuring the ability of models to generalize to superficially novel yet functionally equivalent in-context learning problems. By exploiting the principled statistical framework of Gaussian Process regression, we can further characterize the priors that LLMs have about continuous functions, and the power of post-training methods to steer those inductive biases. Indeed, a key debate in natural language processing is the differential power of token-based reinforcement learning compared to supervised fine-tuning \citep{chu2025sft}.

In this study, we found evidence of strong in-context learning capabilities for novel multivariate function learning problems, confirming existing results \citep{nafar2024learning,requeima2024llm,vacareanu2024words}. The error rates on these problems are surprisingly low, competitive with the empirical GP and far outperforming a data-driven 1-nearest-neighbor algorithm, even for higher-dimensional problems. This suggests that these models are not simply deriving predictions from previously observed demonstrations \citep{vacareanu2024words}.

We also found that LLMs appear to be inductively biased towards rougher continuous functions for lower dimensional problems and smoother continuous functions for higher dimensional problems. This is counter-intuitive in the context of function learning in humans, who appear to show a bias towards smoother, more predictable functions in all regimes \citep{schulz2015assessing}. The exact reason for this phenomenon may either derive from features of the attention mechanism \citep{xie2021explanation} or from properties of the large-scale pre-training data used to train these models \citep{cruz2023reinforcement}.

Both supervised fine-tuning and group-relative policy optimization over a small set of injected low-rank adapter matrices on each layer where able to effectively shift these preferences towards the structure of the training data. Furthermore, we found some evidence pointing in the same direction as \citet{chu2025sft}, suggesting that reinforcement-based post-training (e.g., GRPO) leads to more generalizable behavior than supervised fine-tuning, which tends to memorize the training data.

There are several limitations to this work. First, our inductive bias analysis suffers from a variance inflation problem, meaning that model predictions are always as or more likely under rougher kernels due to larger variances. Future work will explore alternative corrections for this beyond the case described in the supplementary material (Section 8). Second, it is not yet clear how our results generalize from the controlled setting of function learning to more ecologically valid in-context learning settings, such as interacting with a user or iterating over a novel coding task.

\section{Related Work}
In-Context Learning (ICL) in Large Language Models was most prominently discussed by \citet{brown2020language}, who showed that large-scale autoregressive models can learn to solve a wide range of tasks purely from demonstrations at inference time. Subsequent work has sought to reveal the mechanisms underlying this capacity. \citet{xie2021explanation} interpret ICL as implicit Bayesian inference over latent concepts, while \citet{akyurek2022learning} formulate it as an approximation of gradient descent performed during the forward pass \citep{von2023transformers}. Alternative perspectives include structure induction, where transformers recombine latent compositional patterns learned during pre-training to induce task structure \citep{hahn2023theory}. Empirically, larger models appear to display qualitatively different sensitivities to context noise compared to smaller models \citep{wei2023larger}, and prior work has examined whether models can learn specific function classes such as linear and sparse regressions and how they compare to classical estimators \citep{bhattamishra2023understanding,garg2022can}.
\paragraph{Inductive Biases in ICL}
\citet{wei2023larger} investigated feature biases using underspecified prompts, showing that GPT models systematically prefer certain predictive features (e.g., sentiment over lexical markers). \citet{coda2023meta} find that GPT-3 exhibits biases in favour of positive monotonic functions, but that these biases can be shifted through in-context demonstrations. This connects to a long tradition in computational cognitive science, where inductive biases are inferred from point estimates in function learning tasks \citep{griffiths2008modeling,li2023learning,lucas2015rational,schulz2018tutorial,wilson2015human}. Humans themselves appear to favour smoother, more predictable functions \citep{schulz2015assessing}.
\paragraph{Regression and Function Learning with LLMs}
Recent work has examined regression capabilities of LLMs directly. \citet{vacareanu2024words} show that models such as GPT-4 and Claude 3 can outperform classical supervised learners on certain non-linear regression tasks, while \citet{nafar2024learning} demonstrate that performance depends strongly on the presentation of in-context demonstrations. To better characterise predictive behaviour, \citet{requeima2024llm} introduce LLM Processes, eliciting predictive distributions directly from LLMs and comparing them against regression baselines. Several theoretical studies further suggest that transformers may implicitly implement kernel or least-squares regression mechanisms \citep{han2023explaining,zhang2024trained,sun2025context,bai2023transformers}. Our work builds on these insights by evaluating LLMs on controlled function learning tasks and testing how well Gaussian Process (GP) models with appropriate kernels can reproduce their in-context predictions.
\paragraph{Bayesian Perspectives, Calibration, and Uncertainty}
From a Bayesian viewpoint, \citet{zhang2305and} argue that ICL corresponds to approximate Bayesian model averaging over latent functions implemented through attention, while \citet{falck2024context} show that LLMs violate certain Bayesian properties such as the martingale condition. Related work studies reliability and uncertainty estimation: \citet{chen2023calibrating} introduce Sparse Gaussian Process Attention to improve calibration, and \citet{jesson2024estimating} quantify hallucination rates as low-likelihood outputs under an assumed latent model. Rather than modifying model architecture, our approach analyses the natural calibration of LLM outputs on function learning tasks and evaluates how well GP post-hoc modelling captures predictive uncertainty.
\paragraph{Post-Training and Steering Inductive Biases}
Several studies examine how post-training methods can shift inductive biases. Reinforcement learning from human feedback appears to bias models toward extractable features \citep{cruz2023reinforcement}, and reinforcement learning-based post-training may improve generalisation compared to supervised fine-tuning \citep{chu2025sft}. Our work complements these findings by studying how fine-tuning influences inductive biases in controlled function learning settings.

\section{Conclusion}

Any system interacting with an environment inevitably handles continuous functional relationships, from judging the sentiment of a piece of text to forecasting the weather. Although LLMs are often evaluated on discrete tasks, their routine use often requires learning smooth input-output relationships like mapping text to scores, rewards, or making continuous predictions from a few examples. 

Our study casts in-context function learning as a principled test bed for probing the statistical capabilities and inductive biases of large language models. By grounding our analysis in Gaussian process regression, we demonstrate that LLMs can approximate non-parametric regression learners, with performance scaling with model size and approaching the GP lower bound with more demonstrations. However, our inductive bias analysis reveals a consistent tendency toward rougher functions, pointing to a divergence from human expectations.

We further show that post-training with parameter-efficient methods such as supervised fine-tuning and reinforcement learning can shift these biases towards previously unlikely kernels. While supervised fine-tuning exhibits signs of overfitting, reinforcement learning appears to promote more generalizable adjustments. These findings highlight both the promise and limitations of LLMs as function learners, and point to controlled function learning tasks as a powerful framework for dissecting and steering in-context learning mechanisms.

\section*{Acknowledgements}

This project has received funding from the European Research Council (ERC) under the European Union's Horizon Europe research and innovation programme (ERC Starting Grant TACOS). We thank the International Max Planck Research School for Intelligent Systems (IMPRS-IS) for supporting EA.

\clearpage
\bibliographystyle{plainnat}
\bibliography{refs}

\end{document}


%
\runningtitle{In-Context Function Learning in LLMs}

%

\onecolumn
\aistatstitle{In-Context Function Learning in Large Language Models: \\
Supplementary Materials}

\section{A Guideline For Studying Inductive Biases in LLMs}

This section summarises the general, model-agnostic procedure we took for analysing inductive biases in LLMs on continuous function learning tasks. It draws directly from classic methods from cognitive science, where inductive biases are inferred from human judgements \citep{griffiths2008modeling,griffiths2006optimal,lucas2015rational,schulz2018tutorial}. The same framework can be applied to LLMs since they can be understood as conditional predictors receiving structured demonstrations, analogous to the human case.

The procedure consists of four components:

\begin{enumerate}
\item Construct controlled function-learning tasks by sampling target functions from known generative families, mirroring experimental designs in human function-learning research \citep{mcdaniel2005conceptual,schulz2015assessing}. In our study, we used functions sampled from Gaussian Processes with Matérn or Squared Exponential kernels.

\item Evaluate in-context learning curves by measuring prediction error as a function of the number of in-context demonstrations. These learning curves should be compared to principled references that implement interpretable learning prediction strategies. In our case, we used a simple memorisation baseline (1-nearest neighbour) to serve as an upper bound on non-generalising behaviour, and Gaussian-process regression on the same data, as an empirical lower bound under an assumed prior. This allows us to assess whether the model behaves more like a non-parametric regressor \citep{coda2023meta,requeima2024llm} or relies primarily on pattern retrieval \citep{vacareanu2024words}.

\item Perform a likelihood-based inductive bias analysis by considering the LLM’s point predictions given some data under the posterior predictive distribution of a Gaussian process with a specific kernel. Aggregation of these likelihoods across all predictions identifies the kernel under which the LLM’s predictions are most likely. This method follows from research on inferring inferring human inductive biases via predictive likelihoods (Griffiths et al., 2008; Lucas et al., 2015; Li et al., 2023) and connects naturally to work on structured statistical inductive reasoning \citep{tenenbaum2011grow,kemp2009structured}.
\end{enumerate}

Our approach provides a unified testbed for analysing the in-context inductive biases of LLMs, rooted in existing methodologies in computational cognitive science and non-parametric Bayesian statistics. It supports controlled comparisons across model families, training regimes, and post‑training interventions, and offers a principled way to interpret in‑context learning behavior through the lens of functional priors.

\section{Expected 1-Nearest Neighbour Mean Absolute Error}

To compute the 1-NN expected absolute error, we assume $n$ inputs are drawn uniformly as $X \sim \mathcal{U}[0, L]$ with some upper bound, $L$. We fix $i\in\{1,\dots,n\}$ and let $j^\star(i)$ be the index of $x_i$'s nearest
neighbour among $\{x_j\}_{j\neq i}$.

The 1-NN error is given by:

$$\varepsilon_i := y_i - y_{j^\star(i)} = (f(x_i) - f(x_{j^\star(i)})) + (\epsilon_i + \epsilon_{j^\star(i)})$$

Where $\epsilon \sim \mathcal{N}(0, \sigma^2_\epsilon)$ is the noise added to the outputs of the function $f(\cdot)$ drawn from the GP. The error, $\varepsilon_i$ is conditional on the distance, $d_i$, between $x_i$ and $x_{j^\star(i)}$ is defined as $|x_i - x_{j^\star(i)}|$. Assuming a Gaussian process with kernel function $k(\cdot)$, mean 0, and output variance $\sigma^2_f$ (i.e. the value of $k(0)$), the mean of $\varepsilon_i$ is also 0 and the variance is defined as:

$$V(d) = 2\sigma^2_f + 2\sigma^2_\epsilon - 2k(d)$$

Since $\varepsilon$ is normally distributed, the expected \textit{absolute} error conditional on $d$ is therefore:

$$\mathbb{E}[|\varepsilon| \mid d] = \sqrt{\frac 2 \pi} \sqrt{V(d)}$$

To compute the unconditional absolute error $\mathbb{E}[|\varepsilon|]$, we first notice that the probability no member of $\{x_1, ..., x_{n-1}\}$ is within distance $d$ of $x_n$ is $\left(1 - \frac{2d}{L}\right)^{n-1}$, for cases where $x_n$ is at least $d$ away from 0 or $L$. The probability that the closest other input to $x_n$ is at most $d$ is the complement of this probability, and is given by the cumulative density function:

$$F_D(d) = \text{Pr}(D \leq d) = 1 - \left(1 - \frac{2d}{L}\right)^{n-1}$$

where $0 \leq d \leq \frac L 2$. The probability density function is derived by differentiation:

$$f_D(d) = F'_D(d) = \frac{2(n-1)}{L}\left(1 - \frac{2d}{L}\right)^{n-2}$$

with the same support. Therefore, the unconditional expected absolute error is given by:

$$\mathbb{E}[|\varepsilon|] = \sqrt{\frac{2}{\pi}} \int^{L/2}_{0} V(d) \frac{2(n-1)}{L}\left(1 - \frac{2d}{L}\right)^{n-2} \text{d}d$$

In this work, this integral is evaluated numerically using adaptive Gauss-Kronrod quadrature.

\section{Prompting Structure}\label{app:prompting}

We used the following prompt structure for 1-dimensional regression tasks:

\begin{quote}
   \textit{ You are a number predictor. I will give you a number, X, and then you need to predict a new number, Y. There may be noise in the true prediction. Your task is to provide your best estimate for Y. Provide that and only that, without any additional text.\\
    X: }\{\texttt{train input}\}\textit{, Y: }\{\texttt{train output}\} ... \textit{X:} \{\texttt{test input}\}\textit{, Y:}
\end{quote}

where there are between 0 and 49 train input-output pairs as in-context demonstrations.

We used the following prompt structure for 2-, 3-, and 4-dimensional regression tasks:

\begin{quote}
   \textit{ You are a function approximator.  I will give you a set of input variables (X), and then you need to the output value (Y). There may be noise in the true prediction. Your task is to provide your best estimate for Y. Provide that and only that, without any additional text.\\
    X0: }\{\texttt{train input 1}\} ... \textit{, Y: }\{\texttt{train output}\} ... \textit{X0:} \{\texttt{test input 1}\} ... \textit{, Y:}
\end{quote}

where again there are between 0 and 49 train input-output pairs as in-context demonstrations. Note that the model is given the role of a function approximator here, rather than a number predictor.

\newpage

\section{Multi-dimensional Regression Task}

We conducted the inductive bias analysis on the multi-dimensional regression problem. Table \ref{tab:multidim-logliks} below shows the log-likelihoods of the the model predictions under each kernel. For all model sizes, predictions are most likely under the Mat\'{e}rn 1$\frac 1 2$ kernel for 2-dimensional regression tasks, and under the Squared Exponential for 3- and 4-dimensional tasks. Recall that model predictions are most likely under the Mat\'{e}rn $\frac 1 2$ kernel for 1-dimensional tasks. This suggests that these models are inductively biased to rougher functions in low-dimensional space and smoother functions in high-dimensional space.

\begin{table}[h]
    \caption{The total log-likelihood over 19,200 predictions by two Qwen-3 models under different kernels on functions generated either from the Squared Exponential (SE) or the Mat\'{e}rn $\frac 1 2$. The highest log-likelihoods for each dimensionality for each model is shown in bold.} \label{tab:multidim-logliks}
    \begin{center}
    \begin{tabular}{lllccc}
    Parameters & \textbf{Likelihoods Under}  & \textbf{Functions From} & \textbf{2} & \textbf{3} & \textbf{4} \\
    \hline \\
    8B & Mat\'{e}rn $\frac{1}{2}$ & SE &  $-9.81\times10^3$ & $-1.47\times10^4$ & $-1.69\times10^4$ \\
    & & Mat\'{e}rn $\frac{1}{2}$ &  $-9.72\times10^3$ & $-1.47\times10^4$ & $-1.69\times10^4$\\
    & Mat\'{e}rn $1\frac{1}{2}$ & SE &  $\mathbf{-2.05\times10^3}$& $-1.15\times10^4$ & $-1.60\times10^4$ \\
    & & Mat\'{e}rn $\frac{1}{2}$ &  $\mathbf{-1.65\times10^3}$& $-1.14\times10^4$ & $-1.60\times10^4$ \\
    & Mat\'{e}rn $2\frac{1}{2}$ & SE &   $-1.06\times10^4$& $-9.90\times10^3$& $-1.56\times10^4$ \\
    & & Mat\'{e}rn $\frac{1}{2}$ &  $-1.17\times10^4$& $-9.81\times10^3$& $-1.56\times10^4$\\
    & SE & SE &  $-8.56\times10^6$ & $\mathbf{-7.12\times10^3}$& $\mathbf{-1.44\times10^4}$ \\
    & & Mat\'{e}rn $\frac{1}{2}$ &  $-9.26\times10^6$& $\mathbf{-7.02\times10^3}$ & $\mathbf{-1.44\times10^4}$  \\
    \hline
    14B & Mat\'{e}rn $\frac{1}{2}$ & SE & $-9.68\times10^3$  & $-1.47\times10^4$& $-1.68\times10^4$ \\
    & & Mat\'{e}rn $\frac{1}{2}$ & $-9.63\times10^3$ & $-1.46\times10^4$& $-1.68\times10^4$ \\
    & Mat\'{e}rn $1\frac{1}{2}$ & SE & $\mathbf{-1.00\times10^3}$& $-1.15\times10^4$& $-1.59\times10^4$ \\
    & & Mat\'{e}rn $\frac{1}{2}$ & $\mathbf{-8.39\times10^2}$& $-1.14\times10^4$& $-1.59\times10^4$ \\
    & Mat\'{e}rn $2\frac{1}{2}$ & SE & $-6.12\times10^3$ & $-9.87\times10^3$ & $-1.55\times10^4$ \\
    & & Mat\'{e}rn $\frac{1}{2}$ & $-7.68\times10^3$& $-9.79\times10^3$ & $-1.55\times10^4$ \\
    & SE & SE & $-7.65\times10^6$& $\mathbf{-6.92\times10^3}$& $\mathbf{-1.43\times10^4}$ \\
    & & Mat\'{e}rn $\frac{1}{2}$ & $-8.45\times10^6$& $\mathbf{-7.03\times10^3}$& $\mathbf{-1.43\times10^4}$ \\
    \hline
    \end{tabular}
    \end{center}
\end{table}

\section{Lengthscale Comparison}\label{app:lengthscale-priors}

We use the same methodology as in the main paper to determine under which $\ell$ value for the Squared Exponential (SE) kernel are LLM predictions most likely. Table \ref{tab:ls-comparison-se} shows the likelihoods on SE-sampled functions for the 8B and 14B base models before and after post-training with SFT or GRPO. Table \ref{tab:ls-comparison-mat} shows the same for Mat\'{e}rn$\frac 1 2$ sampled functions. We see that LLM predictions are much more likely under kernels with lower $\ell$. After training these models for 2,000 steps on data drawn from the SE  with $\ell=8$, we notice that the likelihood of LLM responses increases substantially for both post-training methods on SE-sampled functions, with that difference being greater for SFT-trained models. However, this pattern reverses for Mat\'{e}rn$\frac 1 2$-sampled functions. The 8B SFT model's predictions are less likely under the SE with $ls = 6,8$, and the 14B SFT model's predictions are less likely under all length scales. Meanwhile, the GRPO models show a comparable likelihood increase across all lengthscales. This result is in favour of the hypothesis that GRPO enables models to learn something generalizable about the functional structure of their training data, while SFT may simply contribute to memorization of those data.

\begin{table}[h]
    \caption{The average likelihood per prediction under the Squared Exponential with different length scales, $\ell$, on functions drawn from the Squared Exponential, for the 8B and 14B models before and after fine-tuning with either SFT or GRPO. Both models were trained with data drawn from the Squared Exponential with $\ell=8$ for 2,000 steps. $\Delta$ shows the difference (SFT/GRPO - Base).} \label{tab:ls-comparison-se}
    \begin{center}
    \begin{tabular}{lrrrrrr}
    \textbf{Size}  & \textbf{$\ell$} & \textbf{Base} & \textbf{SFT} & $\Delta$ &  \textbf{GRPO} & $\Delta$ \\
    \hline \\
       8B & 1 & $-1.95 \times10^2$ & $-2.76\times10^1$ & $+1.67\times10^2$ & $-6.44\times10^1$ & $+1.31\times10^2$\\
          & 2 & $-1.28 \times10^3$ & $-3.52\times10^2$ & $+9.28\times10^2$ & $-6.30\times10^2$ & $+6.50\times10^2$\\
          & 4 & $-3.63 \times10^3$ & $-1.76\times10^3$ & $+1.87\times10^3$ & $-2.21\times10^3$
          & $+1.42\times10^3$\\
          & 6 & $-5.27 \times10^3$ & $-2.95\times10^3$ & $+2.32\times10^3$ & $-3.27\times10^3$ & $+2.00\times10^3$\\
          & 8 & $-6.39 \times10^3$ & $-3.83\times10^3$ & $+2.56\times10^3$ & $-3.99\times10^3$ & $+2.40\times10^3$\\
    \hline \\
      14B & 1 & $-3.95 \times10^1$ & $-1.35\times10^1$ & $+2.60\times10^1$ & $-2.32\times10^1$ & $+1.63 \times10^1$ \\
          & 2 & $-5.03 \times10^2$ & $-2.24\times10^2$ & $+2.79\times10^2$ & $-3.82\times10^2$ & $+1.21 \times10^2$ \\
          & 4 & $-1.91 \times10^3$ & $-1.38\times10^3$ & $+5.30\times10^2$ & $-1.66\times10^3$ & $+1.74 \times10^3$ \\
          & 6 & $-2.87 \times10^3$ & $-2.66\times10^3$ & $+2.10\times10^2$ & $-2.64\times10^3$ & $+2.30 \times10^2$ \\
          & 8 & $-3.50 \times10^3$ & $-3.69\times10^3$ & $+1.90\times10^2$ & $-3.32\times10^3$ & $+1.80 \times10^2$ \\
    \hline
    \end{tabular}
    \end{center}
\end{table}

\begin{table}[h]
    \caption{The average likelihood per prediction under the Squared Exponential with different length scales, $\ell$, on functions drawn from the Mat\'{e}rn $\frac 1 2$, for the 8B and 14B models before and after fine-tuning with either SFT or GRPO. Both models were trained with data drawn from the Squared Exponential with $\ell=8$ for 2,000 steps. $\Delta$ shows the difference (SFT/GRPO - Base).} \label{tab:ls-comparison-mat}
    \begin{center}
    \begin{tabular}{lrrrrrr}
    \textbf{Size}  & \textbf{$\ell$} & \textbf{Base} & \textbf{SFT} & $\Delta$ &  \textbf{GRPO} & $\Delta$ \\
    \hline \\
       8B & 1 & $-4.25 \times10^2$ & $-3.93 \times10^2$ & $+3.10\times10^1$ & $-2.47 \times10^2$ & $+1.78 \times10^2$ \\
          & 2 & $-2.01 \times10^3$ & $-1.68 \times10^3$ & $+3.29\times10^2$ & $-1.26 \times10^3$ & $+7.50 \times10^2$ \\
          & 4 & $-3.93 \times10^3$ & $-3.62 \times10^3$ & $+3.10\times10^2$ & $-2.58 \times10^3$ & $+1.35 \times10^3$ \\
          & 6 & $-4.69 \times10^3$ & $-4.93 \times10^3$ & $-2.40\times10^2$ & $-3.31 \times10^3$ & $+1.38 \times10^3$ \\
          & 8 & $-5.03 \times10^3$ & $-5.98 \times10^3$ & $-9.50\times10^2$ & $-3.78 \times10^3$ & $+1.25 \times10^3$ \\
    \hline \\
      14B & 1 & $-2.20 \times10^2$ & $-4.19 \times10^2$ & $-1.99 \times10^2$ & $-1.78 \times10^2$ & $+4.20 \times10^1$ \\
          & 2 & $-1.14 \times10^3$ & $-1.67 \times10^3$ & $-5.30 \times10^2$ & $-9.85 \times10^2$ & $+1.55 \times10^2$ \\
          & 4 & $-2.32 \times10^3$ & $-3.43 \times10^3$ & $-1.11 \times10^3$ & $-2.11 \times10^3$ & $+2.10 \times10^2$ \\
          & 6 & $-2.93 \times10^3$ & $-4.72 \times10^3$ & $-1.79 \times10^3$ & $-2.77 \times10^3$ & $+1.60 \times10^2$ \\
          & 8 & $-3.31 \times10^3$ & $-5.75 \times10^3$ & $-2.44 \times10^3$ & $-3.22 \times10^3$ & $+9.00 \times10^1$ \\
          \hline
    \end{tabular}
    \end{center}
\end{table}
\newpage
\section{Further Experimental Conditions}\label{app:extra-exps}

\subsection{Training On Data With A Different Length-Scale}

\begin{figure*}[h]
\includegraphics[width=\textwidth]{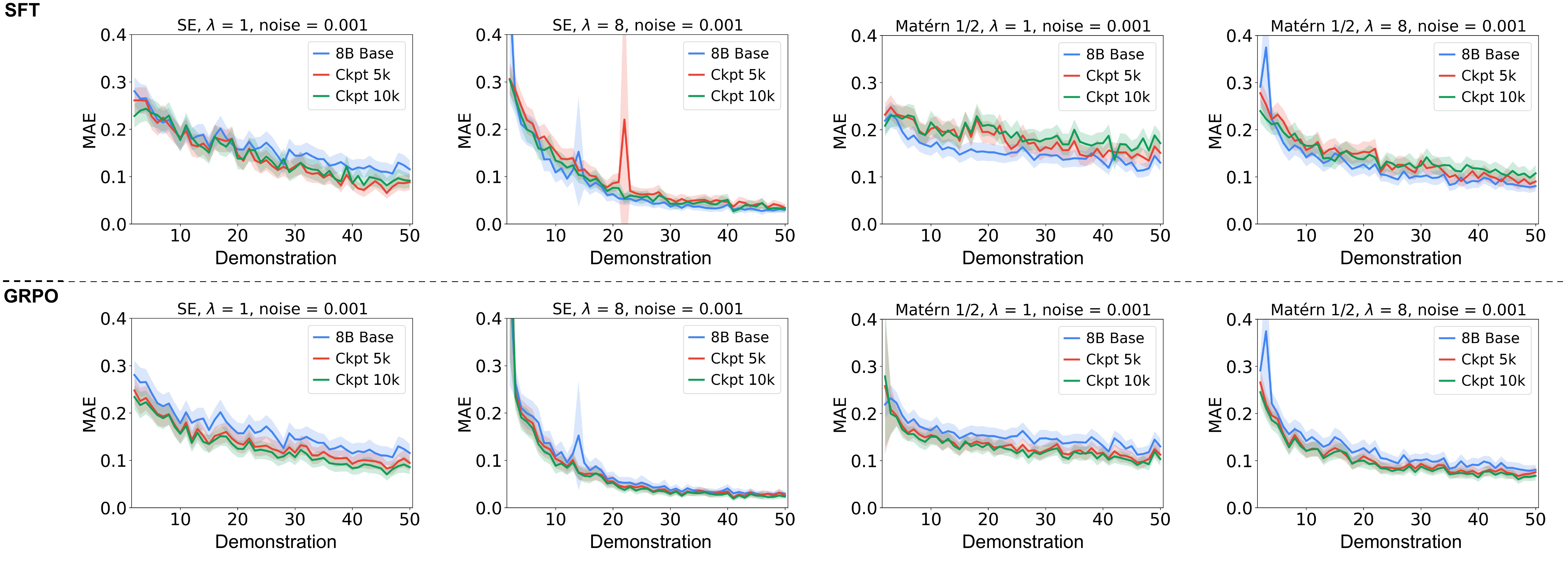}
\caption{Post training learning curves with $\ell=1$ training regime. Top row: Effects of Supervised Fine-Tuning. 8B base model compared to two checkpoints during training (5k, 10k steps). Bottom row: GRPO progression. 8B base model compared to two checkpoints (5k, 10k steps). All empirical learning curves are shown with 95\% bootstrapped confidence intervals.}\label{fig:ls-1-learning-curves}
\end{figure*}

\begin{figure*}[h]
\includegraphics[width=\textwidth]{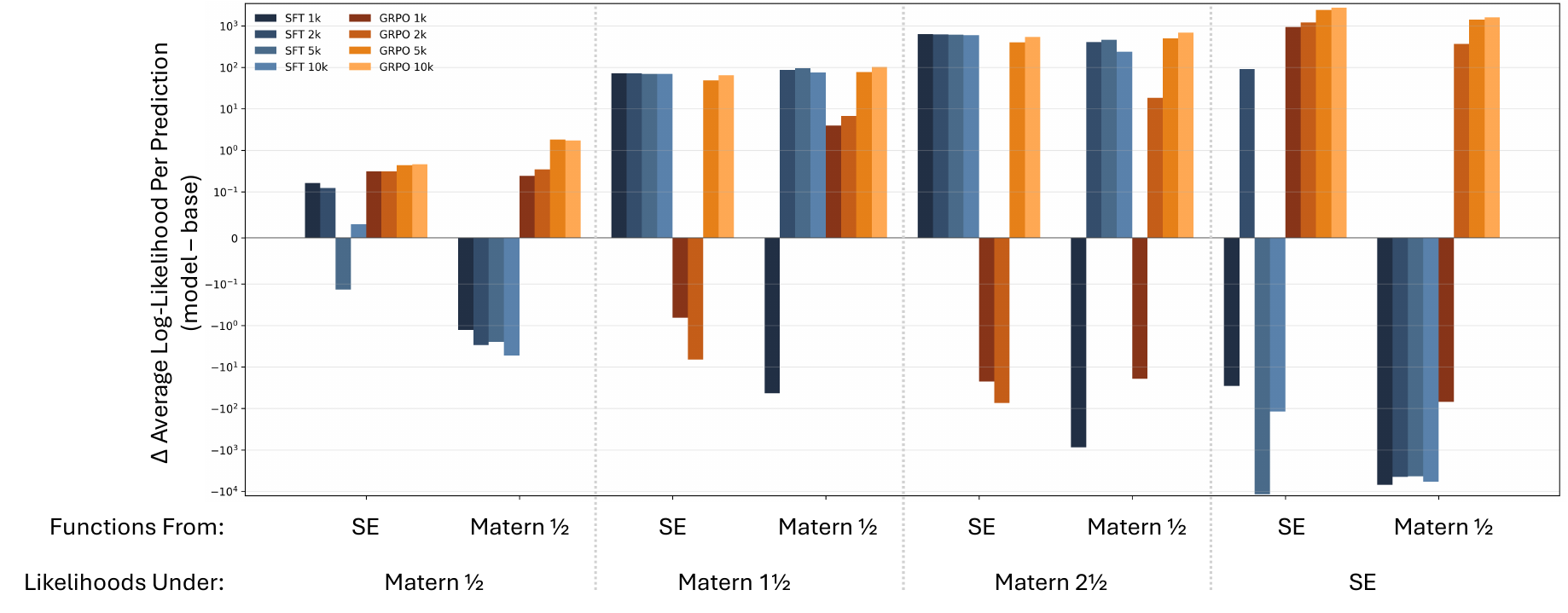}
\caption{Inductive bias analysis compared to the base model for Qwen-3-8B for three checkpoints (1k, 2k, 5k, 10k steps), with training data on the squared exponential with $\ell=1$. Average likelihood per prediction shown on a symlog scale.}\label{fig:ls-1-inductive-biases}
\end{figure*}

We provide an alternative training regime in which the models are trained on data drawn from the Squared Exponential with $\ell=1$ rather than $\ell=8$. We find a broadly similar pattern with respect to learning curves (Figure \ref{fig:ls-1-learning-curves}) and inductive bias analyses after post-training as shown in Figure \ref{fig:ls-1-inductive-biases}.

\newpage
\subsection{Training GRPO With A Different Reward Function}

\begin{figure*}[h]
\includegraphics[width=\textwidth]{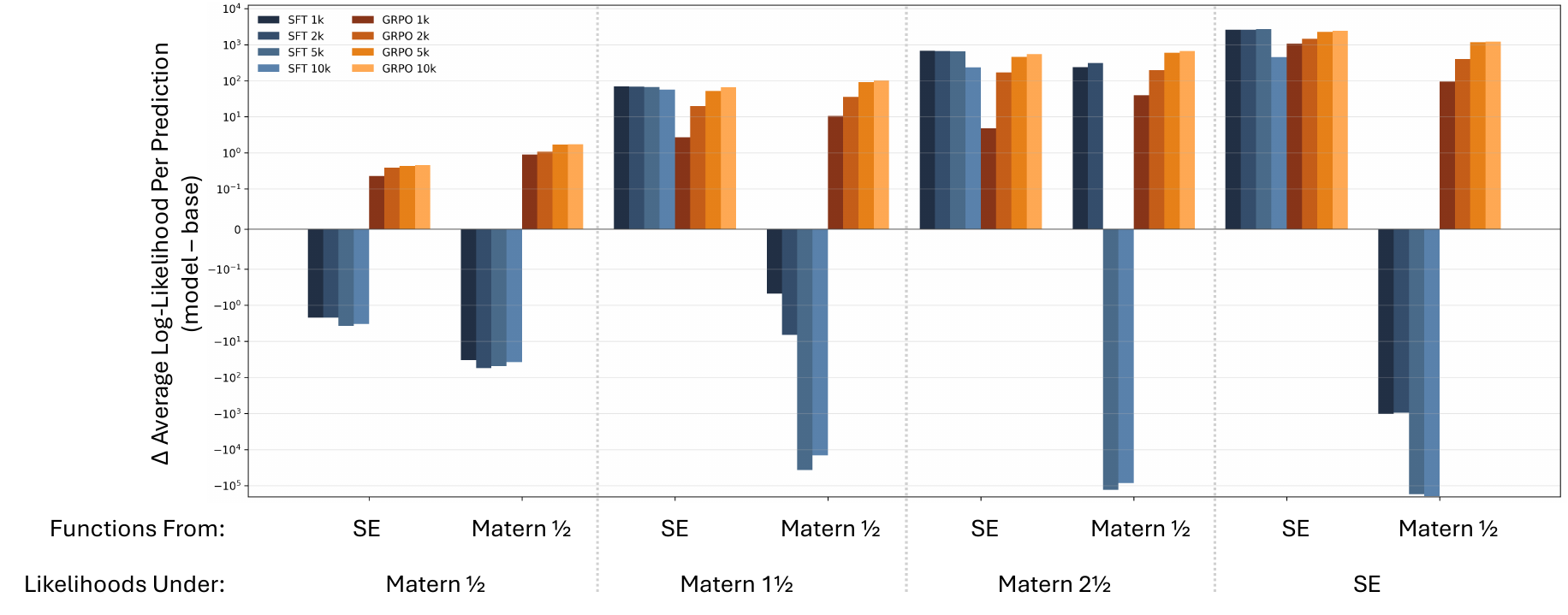}
\caption{Inductive bias analysis compared to the base model for the 8B model for three checkpoints (1k, 2k, 5k, 10k steps), with training data on the squared exponential with $\ell=8$. GRPO was trained directly with a log-likelihood-based reward function. Average likelihood per prediction shown on a symlog scale.}\label{fig:log-lik-grpo-inductive-biases}
\end{figure*}

Reinforcement learning-based post-training is flexible in that several different reward functions can be specified. We use two: negative absolute error and log-likelihood under the SE ($\ell=8$). In both cases, the model's prediction is taken to be the sequentially last float in their completion. For the negative absolute error, reward is capped at a minimum of $-10$ for parseable responses, and is $-11$ for any completions that do not contain a parseable float. For the log-likelihood error, we cap the reward at a minimum of $-999$, and set the reward for unparseable responses at $-1000$. We find no interpretative differences between these two reward functions. Figure \ref{fig:log-lik-grpo-inductive-biases} presents the difference between the post-trained models and the base model in terms of the log-likelihood of their predictions under four different kernels for functions sampled from two different kernels. Once again, we find that GRPO robustly shifts model inductive biases towards smoother kernels for all data regimes, while SFT only shifts these biases under identical regimes to the training data (i.e., the Squared Exponential). 

\subsection{On The Choice Of Kernel Families}

In this work, we focus on the Matérn and squared-exponential (SE) families with varying length scales. Together, these kernels span a broad and interpretable range of function classes from comparatively rough to very smooth sample paths. They also support a simple, likelihood-based comparison that remains easy to interpret across conditions. 

Periodic and compositional kernels are also an important and interesting direction. However, we treat them as beyond the scope of the present study for two reasons. First, exactly periodic kernels exhibit qualitatively different behavior from Matérn/SE in which sample paths from a truly periodic kernel lie in a function class that is poorly captured by standard non-periodic kernels. This would make our likelihood comparisons less interpretable in the framework adopted here. Second, compositional kernels substantially expand the model space and introduce many additional hyperparameters, complicating the analysis and its interpretability. For these reasons, we begin with a compact stationary kernel family and leave a systematic investigation of periodic and compositional kernels to future work.

\newpage

\section{Experiments With Further Models}

Alongside the three Qwen-3 models discussed in the main paper, we also evaluated three members of the Llama family (Llama-3.2-3B-Instruct, Llama-3.1-8B-Instruct, Llama-3.1-70B-Instruct; \citeauthor{grattafiori2024llama}, \citeyear{grattafiori2024llama}), three members of the Gemma-3 family (4B, 12B, and 27B; \citeauthor{team2024gemma}, \citeyear{team2024gemma}), and two members of the Mistral family (Mistral-7B-Instruct-v0.3 and Mistral-Small-24B-Instruct-2501; \citeauthor{qiang2023mistral7b}, \citeyear{qiang2023mistral7b}), where B denotes the parameter size in billions and all models are 4-bit quantized.

\begin{figure*}[t!]
\includegraphics[width=1\textwidth]{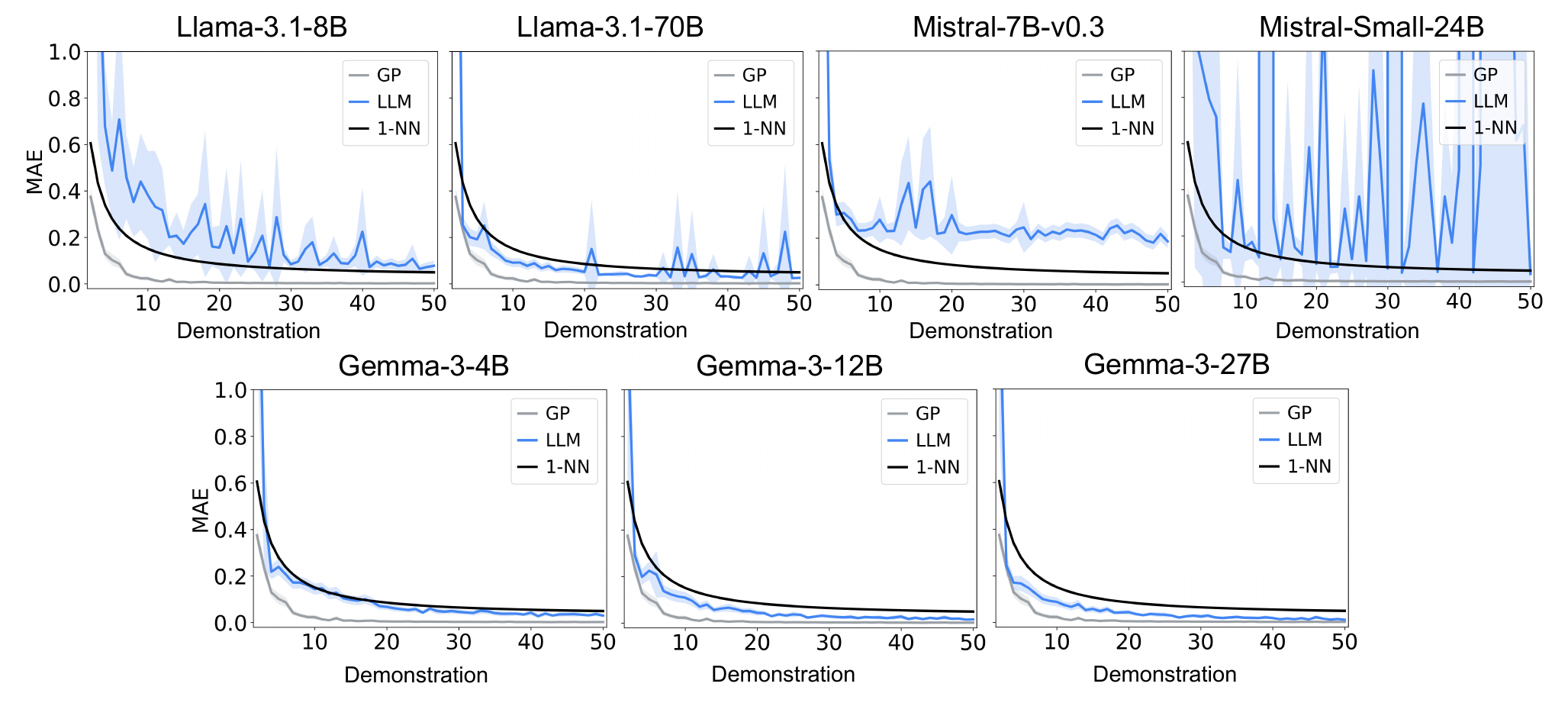}
\caption{Learning curve analysis on 1-dimensional functions. The mean absolute error after $n$ demonstrations by function type. Models from the Llama, Mistral, and Gemma families were tested. Llama-3.2-3B is omitted because error rates were consistently well above 1. All LLM and GP learning curves are shown with 95\% bootstrapped confidence intervals.}\label{fig:learning-curves-others}
\end{figure*}

Figure \ref{fig:learning-curves-others} shows the learning curves for all of these models except for Llama-3.2-3B, which had absolute errors well beyond the observed scale of $y$, suggesting that this model was unable to calibrate to the regression task. Within the Llama family, larger models were better able to perform the task, with only Llama-3.1-70B able to occasionally outperform the 1-NN baseline. The Gemma family were altogether better at the task, with a slight but noticeable effect for model size. The Mistral family performed very poorly, and the larger 24B model appears to be much worse than the smaller 7B model.

\begin{figure*}[t!]
\includegraphics[width=1\textwidth]{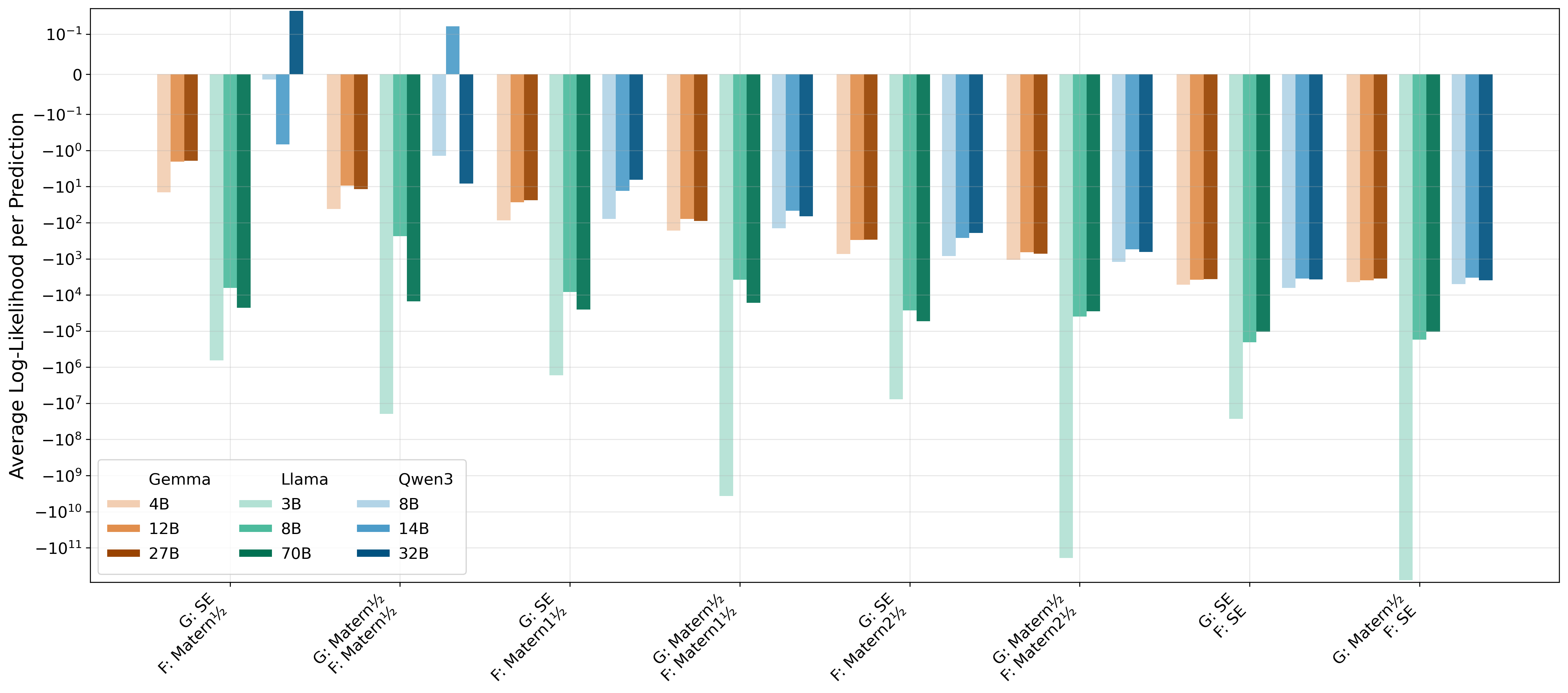}
\caption{Inductive bias analysis of the base models from three model families: Gemma, Llama, and Qwen-3. The average log likelihood per prediction is computed under GPs with four different kernels ($\ell=8$), on 1-dimensional data drawn from either the Squared Exponential or the Mat\'{e}rn $\frac 1 2$. These likelihoods are presented on a symmetric log scale. LLM predictions for all model families sizes are more likely under kernels with lower $\nu$, i.e., those that describe rougher, less predictable functions.}\label{fig:inductive-bias-all-models}
\end{figure*}

We then conducted our inductive bias analysis on all these models, as shown in Figure \ref{fig:inductive-bias-all-models}. We find that, for all model sizes and function families (SE, Mat\'{e}rn $\frac 1 2$), the LLMs' predictions are most likely under the Mat\'{e}rn $\frac 1 2$ and least likely under the SE for 1-dimensional functions. Indeed, we see a general trend that as $\nu \rightarrow \infty$, LLM predictions become less likely. Predictions from the mistral model were so unlikely under all kernels that we have omitted them from Figure \ref{fig:inductive-bias-all-models}.

\begin{figure*}[h]
\includegraphics[width=\textwidth]{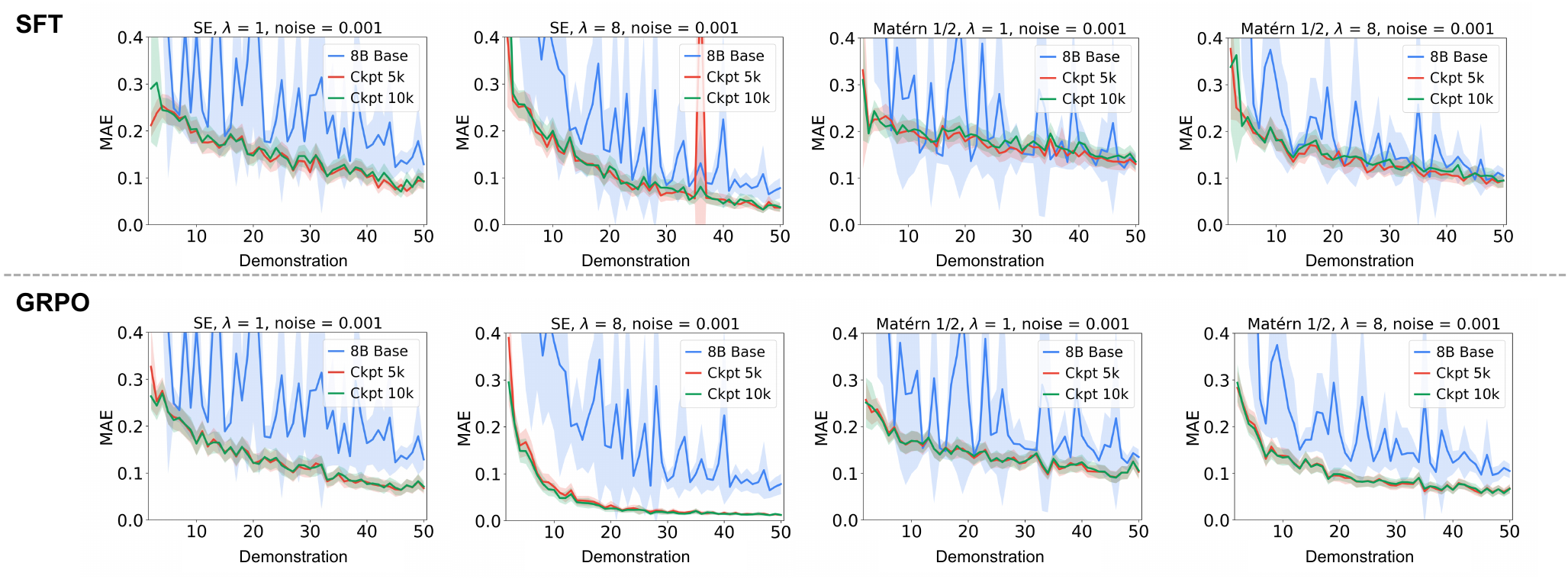}
\caption{Post training learning curves with $\ell=8$ training regime on Llama-3.1-8B. Top row: Effects of Supervised Fine-Tuning. 8B base model compared to two checkpoints during training (5k, 10k steps). Bottom row: GRPO progression. 8B base model compared to two checkpoints (5k, 10k steps). All empirical learning curves are shown with 95\% bootstrapped confidence intervals.}\label{fig:ls-8-learning-curves-llama}
\end{figure*}

\begin{figure*}[h]
\includegraphics[width=\textwidth]{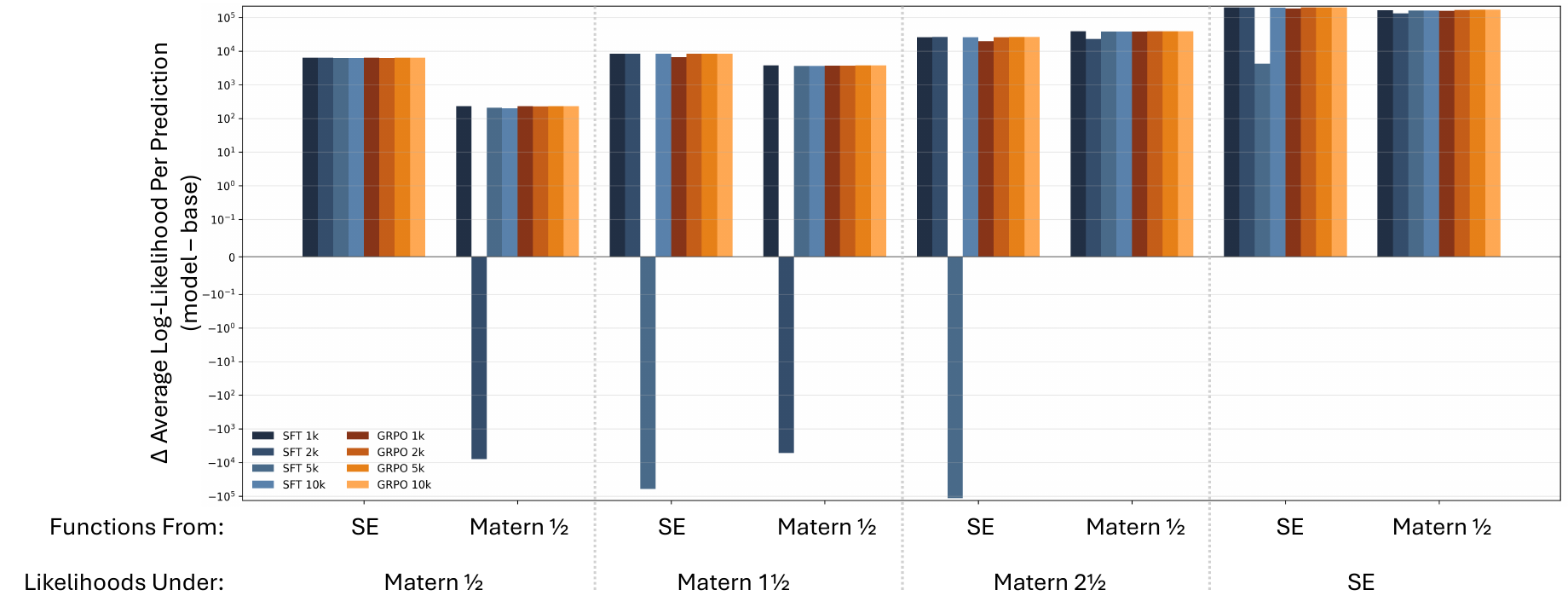}
\caption{Inductive bias analysis compared to the base model for the Llama-3.1-8B for three checkpoints (1k, 2k, 5k, 10k steps) after training on Squared Exponential functions with $\ell=1$. Average likelihood per prediction shown on a symlog scale.}\label{fig:inductive-bias-peft-llama}
\end{figure*}

We repeated our post-training experiments with Llama-3.1-8B, which had a similar pattern to Qwen-3 insofar as its predictions are most likely under the Mat\'{e}rn $\frac 1 2$ kernel and least likely under the Squared Exponential. We found that post-training heavily reduces the noise in the learning curves observed in the base model (Figure \ref{fig:ls-8-learning-curves-llama}). Both SFT and GRPO lead to model predictions being more likely under all kernels (Figure \ref{fig:inductive-bias-peft-llama}), likely because they both lead to a reduction in prediction noise, i.e., they both improve model calibration. In contrast, Qwen-3 already appears to be well-calibrated before any fine-tuning.

\section{Variance Inflation Correction}\label{app:var-inf-cor}

We note that our inductive bias analysis is threatened by what we call a variance inflation effect. Mat\'{e}rn$\frac 1 2$ kernels with lower $\nu$ (for which the Squared Exponential is equivalent to $\nu \rightarrow \infty$) have higher posterior predictive variances. Thus, without adjustment, point predictions are as or more log-likely under kernels with lower $\nu$ compared to those with higher $\nu$.

\begin{figure*}[h]
\includegraphics[width=\textwidth]{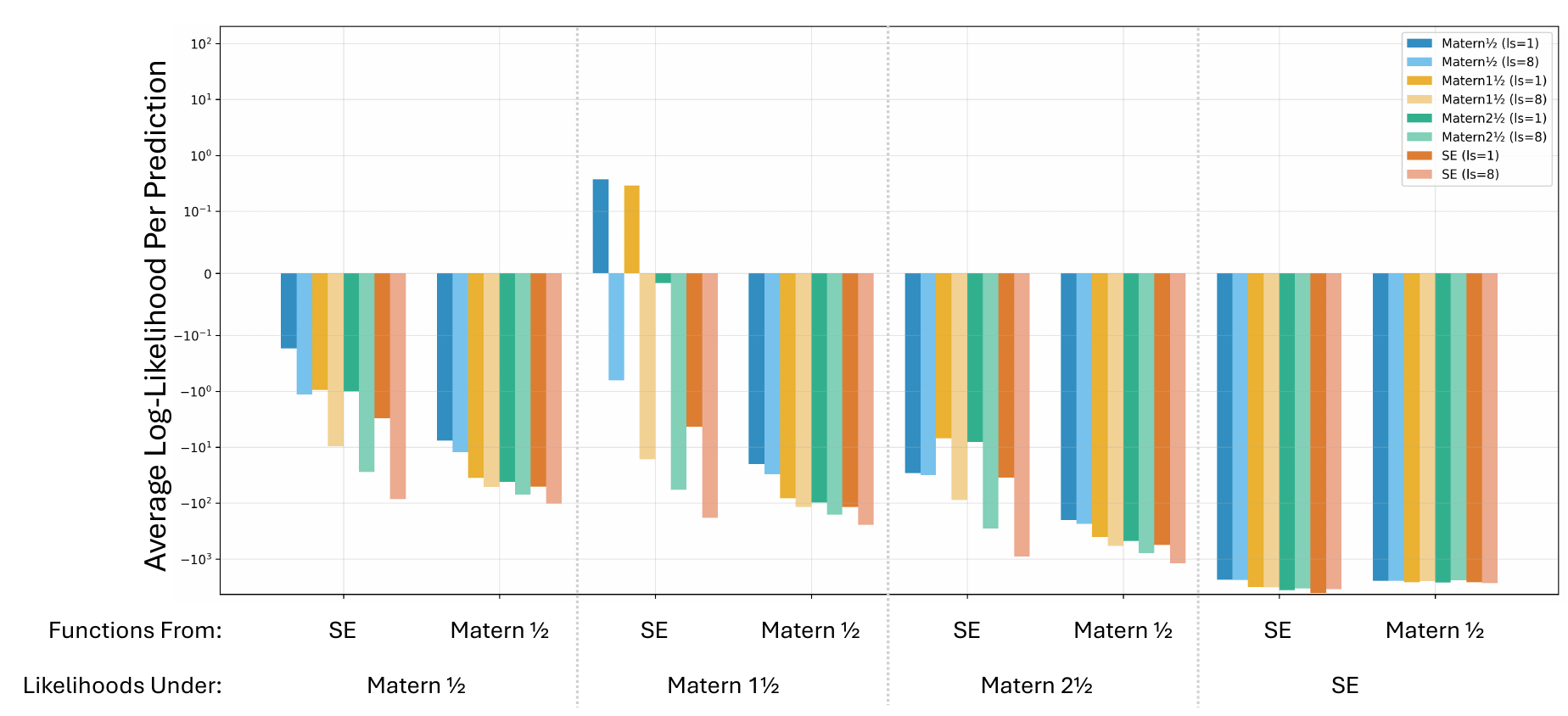}
\caption{Inductive bias analysis on GP predictors on functions generated by two different kernels, under GPs with different kernels. We see that unadjusted likelihoods do not straightforwardly recover the ground truth kernel of the model making the prediction.}\label{fig:gp-inductive-biases}
\end{figure*}

To verify this, we repeated our inductive bias analysis but used Gaussian Processes as our point predictors instead of Large Language Models. Figure \ref{fig:gp-inductive-biases} presents the results of this analysis. As can be seen, the inductive bias analysis does not straightforwardly recover the kernel of the point-predicting GP. All predictions are generally more likely under kernels with lower $\nu$ than kernels with higher $\nu$.

In an attempt to correct for this, we introduce a variance correction. Rather than assuming that point predictions are true draws from a GP posterior predictive ($\hat{y} \sim \mathcal{N}(\mu_i, \sigma_i^2)$ for GP mean and variance $\mu_i, \sigma_i^2$), we instead assume that they are noisy draws with additional variance, $\tau^2$ (i.e., $\hat{y} \sim \mathcal{N}(\mu_i, \sigma_i^2+\tau^2)$).

Since we do not know $\tau^2$ \textit{a priori}, we learn it from the data. For each point prediction under each kernel, we collect all the residuals, $\textbf{r}$, as the difference between the model's prediction, $\hat{y}_i$, and the posterior predictive mean, $\mu_i$, as well as the posterior predictive variances,  $\textbf{v}$. The total log-likelihood of $\tau^2$ across all prediction-kernel pairs is given by:

$$\mathcal{L}(\tau^2) = \sum_i\text{log}~\mathcal{N}(r_i|0,v_i+\tau^2) = - \frac 1 2 \sum_i\left[\text{log}(2\pi(v_i + \tau^2))+\frac{r_i^2}{v_i + \tau^2}\right]$$

We minimize the negative log-likelihood with respect to $\text{log}(\tau^2)$ to enforce $\tau^2>0$, which is a convex optimization problem. This essentially produces a pooled variance across all kernels which serves to reduce the variance inflation effect. Now, the inductive bias analysis more accurately measures the degree to which LLM predictions match the posterior predictive means of each kernel.

\begin{figure*}[!h]
\includegraphics[width=\textwidth]{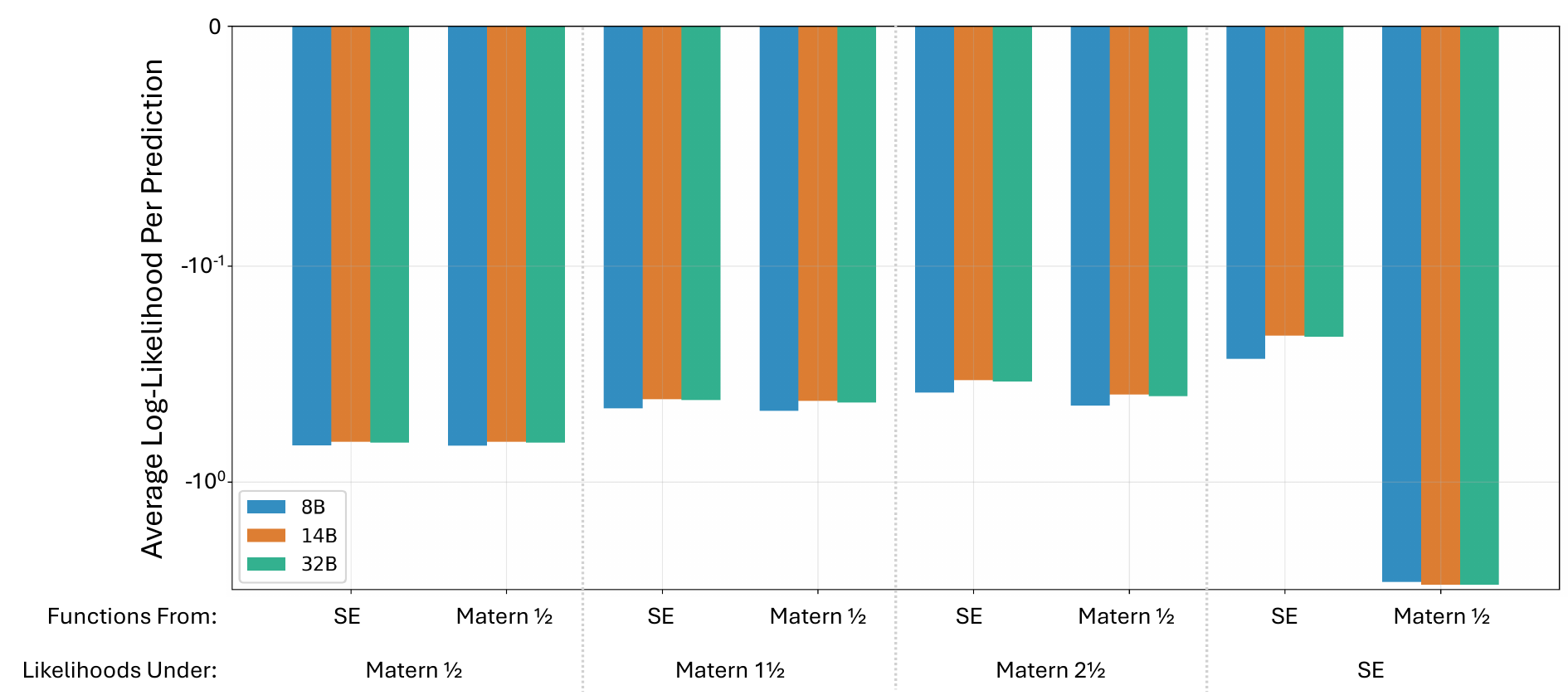}
\caption{Inductive bias analysis of the base models (8B, 14B, 32B) with $\tau^2$-adjustment. Average likelihood per prediction shown on a symlog scale.}\label{fig:tau2-adjustment-base}
\end{figure*}

We use this $\tau^2$-adjustment for our inductive bias analysis, which is presented in Figure \ref{fig:tau2-adjustment-base}. These results show that the base LLM predictions are more likely under smoother kernels, although this partly depends on the structure of the data being administered.

\clearpage
\bibliographystyle{plainnat}
\bibliography{refs}